\documentclass[times,twocolumn,final]{elsarticle}

\usepackage{medima}
\usepackage{framed,multirow}
\usepackage{amssymb}
\usepackage{latexsym}
\usepackage{xcolor}
\usepackage{lineno,hyperref}
\modulolinenumbers[5]
\usepackage{amsmath}
\usepackage{verbatim}
\usepackage{multirow}
\usepackage{color,graphicx}
\usepackage{arydshln}
\usepackage{hyperref}
\usepackage{url}
\usepackage{footnote}
\usepackage{bm}
\usepackage{cite}
\usepackage{subfigure}
\usepackage{verbatim}
\usepackage{ulem}
\newcommand{\tabincell}[2]{\begin{tabular}{@{}#1@{}}#2\end{tabular}}

\providecommand{\Leireftb}[1]{Table~\ref{#1}}
\providecommand{\Leireffig}[1]{Fig.~\ref{#1}}

\providecommand{\leicolor}[2]{\textcolor{blue}{\textit{#2}}}

\providecommand{\reviewerone}[2]{\textcolor{black}{#2}}
\providecommand{\reviewertwo}[2]{\textcolor{black}{#2}}
\definecolor{Mycolor1}{HTML}{036564}
\definecolor{Mycolor2}{HTML}{DE7D2C}
\providecommand{\reviewerthree}[2]{\textcolor{black}{#2}}
\providecommand{\reviewerfour}[2]{\textcolor{black}{#2}}

\usepackage{enumitem}

\journal{Medical Image Analysis}

\begin{document}

\verso{Lei Li \textit{et~al.}}

\begin{frontmatter}

\title{Medical Image Analysis on Left Atrial LGE MRI for Atrial Fibrillation Studies: A Review}  

\author[label1,label2,label3]{Lei Li} 
\author[label3,label4]{Veronika A. Zimmer} 
\author[label3,label4,label5]{Julia A. Schnabel} 
\author[label1]{Xiahai Zhuang*} 
\ead[url]{zxh@fudan.edu.cn}

\address[label1]{School of Data Science, Fudan University, Shanghai, China}
\address[label2]{School of Biomedical Engineering, Shanghai Jiao Tong University, Shanghai, China}
\address[label3]{School of Biomedical Engineering and Imaging Sciences, King’s College London, London, UK}
\address[label4]{Department of Informatics, Technical University of Munich, Germany}
\address[label5]{Helmholtz Center Munich, Germany}

\received{16 June 2021}
\accepted{10 Jan 2022}

\begin{abstract}
Late gadolinium enhancement magnetic resonance imaging (LGE MRI) is commonly used to visualize and quantify left atrial (LA) scars. 
The position and extent of LA scars provide important information on the pathophysiology and progression of atrial fibrillation (AF).
Hence, LA LGE MRI computing and analysis are essential for computer-assisted diagnosis and treatment stratification of AF patients. 
Since manual delineations can be time-consuming and subject to intra- and inter-expert variability, automating this computing is highly desired, which nevertheless is still challenging and under-researched.

This paper aims to provide a systematic review on computing methods for LA cavity, wall, scar, and ablation gap segmentation and quantification from LGE MRI, and the related literature for AF studies. 
Specifically, we first summarize AF-related imaging techniques, particularly LGE MRI. 
Then, we review the methodologies of the four computing tasks in detail and summarize the validation strategies applied in each task \reviewerfour{}{as well as state-of-the-art results on public datasets}. 
Finally, the possible future developments are outlined, with a brief survey on the potential clinical applications of the aforementioned methods. 
The review indicates that the research into this topic is still in the early stages. 
Although several methods have been proposed, especially for the LA cavity segmentation, there is still a large scope for further algorithmic developments due to performance issues related to the high variability of enhancement appearance and differences in image acquisition.

\end{abstract}


\end{frontmatter}


\section{Introduction}

\subsection{Clinical goals}
\begin{figure}[t]\center
    \includegraphics[width=0.35\textwidth]{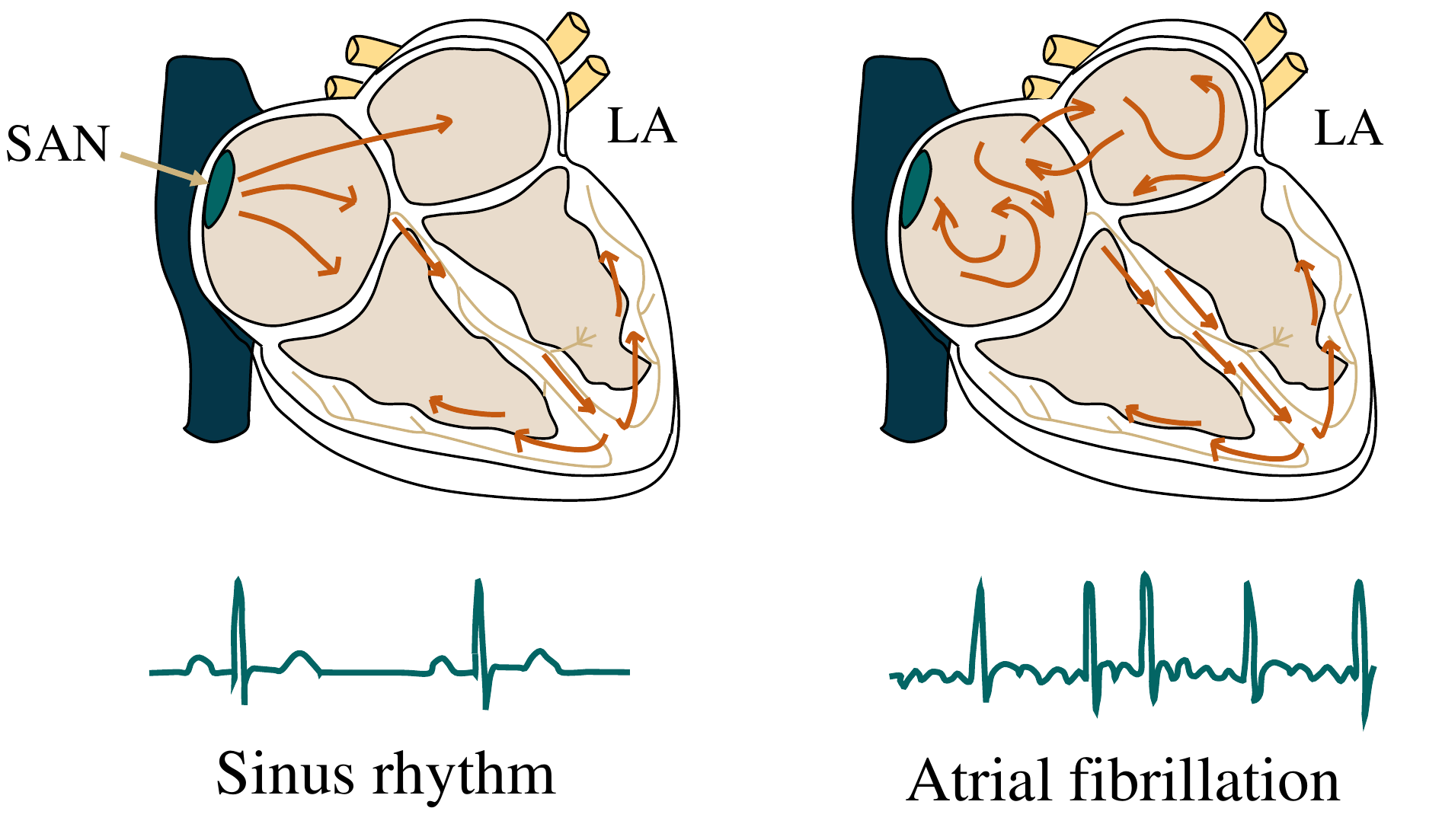}\\[-2ex]
   \caption{\reviewerone{}{The electrical activities of the left atrium (LA) in sinus rhythm and atrial fibrillation (AF), respectively. 
   The sinoatrial node (SAN) produces an electrical impulse, which is regular in the sinus rhythm and can be overwhelmed by disorganized electrical waves, usually originating from the pulmonary veins.}
   }
\label{fig:intro:AF}
\end{figure}

Atrial fibrillation (AF) is the most common cardiac arrhythmia encountered in the clinic, occurring in up to 2\% of the population and rising in prevalence along with advancing age \citep{journal/cir/chugh2014}.
\Leireffig{fig:intro:AF} presents a comparison of sinus rhythm and AF.
One can see that there are chaotic electrical signals in the atrium of AF patients compared to sinus rhythm, resulting in a rapid and irregular heart rhythm.
Radiofrequency catheter ablation via pulmonary vein isolation (PVI) is a promising procedure for treating AF, especially for paroxysmal AF patients \citep{journal/Europace/calkins2007}.
The left atrium (LA) is a crucial structure in the pathophysiology of AF, and the observation of LA remodeling can be important for the initial evaluation of AF \citep{journal/EHJ/tops2010}.
Besides, structural changes in the LA wall (especially changes in the wall thickness) are known to occur in AF patients \citep{journal/MedIA/karim2018}.
The wall thickness can be used to predict the response to invasive treatment of AF and has the potential for improving the safety of AF ablation \citep{journal/EP/whitaker2016}.
The wall thickness is also important to measure the transmurality of scars which is related to the AF recurrence \citep{journal/CAE/ranjan2011}.
The success of AF treatments is highly related to the formation of a contiguous scar completely encircling the veins \citep{journal/CAE/ranjan2011}. 
Unfortunately, the encircling lesion is often incomplete with a combination of ablation scars and gaps of healthy tissue \citep{journal/Europace/miller2012}.
Therefore, the extent and distribution of both scars and gaps are important information for AF patient selection \citep{journal/JCE/akoum2011}, diagnosis prediction \citep{journal/CAE/arujuna2012}, and treatment stratification \citep{journal/EP/njoku2018}.
For example, patients were divided into four grades according to their degrees of fibrosis (refers to preexisting scars) in \citet{journal/JCE/akoum2011}, shown in \Leireftb{tb:intro:utah grade}.
Based on the scoring, various therapeutic strategies were suggested by electrophysiologists.

\begin{table} [t] \center
    \caption{AF patient classification that depends on fibrosis extent \citep{journal/JCE/akoum2011}. 
     }
\label{tb:intro:utah grade}
{\small
\begin{tabular}{l| lll}
\hline
Utah grade     & Percentage  & Success rate &  AF recurrence\\
\hline
Utah 1 (minimal)         & $\leq 5\%$    & 100\%  & 0\\
Utah 2 (mild)            & 5$\sim$20\%   & 81.8\% & 28\%\\
Utah 3 (moderate)        & 20$\sim$35\%  & 62.5\% & 35\%\\
Utah 4 (extensive)       & $\geq 35\%$   & 0      & 56\%\\
\hline
\end{tabular} }\\
\end{table}

\begin{figure*}[t]\center
    \includegraphics[width=0.78\textwidth]{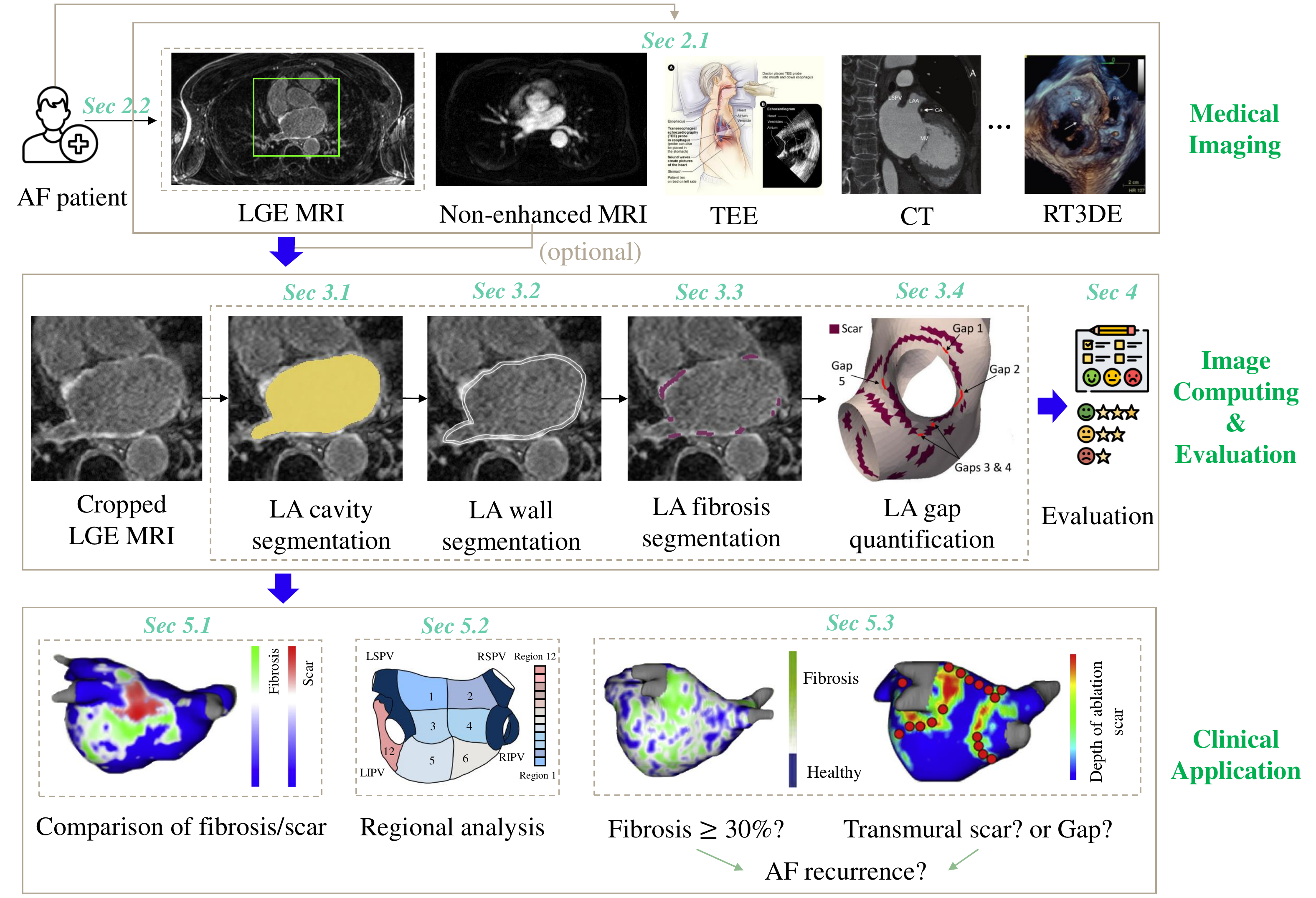}
   \caption{The pipeline of LA image computing for AF studies and the structure of this review. 
   Top row: common image modalities for AF treatments, such as late gadolinium enhancement magnetic resonance imaging (LGE MRI), non-enhanced MRI, transesophageal echocardiography (TEE) (image adapted from \href{https://stanfordhealthcare.org/medical-tests/t/transesophageal-echocardiogram.html}{\textit{Stanford HEALTH CARE}}), CT, and real-time 3D echocardiography (RT3DE) (images adapted from \citet{journal/CY/regazzoli2015} \reviewerone{}{with permission});
   Middle row: computation and evaluation steps for LA analysis reviewed in this study (images adapted from \citet{journal/MedIA/li2021,journal/MedIA/nunez2019} \reviewerone{}{with permission});
   Bottom row: possible clinical applications (images adapted from \citet{journal/JACC/siebermair2017} \reviewerone{}{with permission}).
   }
\label{fig:intro:review_structure}
\end{figure*}

Recently, late gadolinium enhancement magnetic resonance imaging (LGE MRI) has evolved as a tool for defining the extent of fibrosis/ scars and visualizing the ablation gaps \citep{journal/JACC/siebermair2017,journal/MedIA/li2020,journal/MedIA/nunez2019}.
\textit{Therefore, it is crucial to develop techniques for the four progressive tasks, i.e., (1) LA cavity segmentation, (2) LA wall segmentation together with wall thickness measurement, (3) scar segmentation and quantification, and (4) ablation gap localization from LGE MRI.}
\Leireffig{fig:intro:review_structure} provides the clinical pipeline for AF ablation procedures, where the role of LGE MRI is highlighted and the four closely related tasks of clinical interests are presented, followed by several related clinical applications.

\subsection{\reviewerfour{}{Challenges of LA LGE MRI computing}} \label{challenges}
\begin{figure*}[t]\center
    \subfigure[] {\includegraphics[width=1\textwidth]{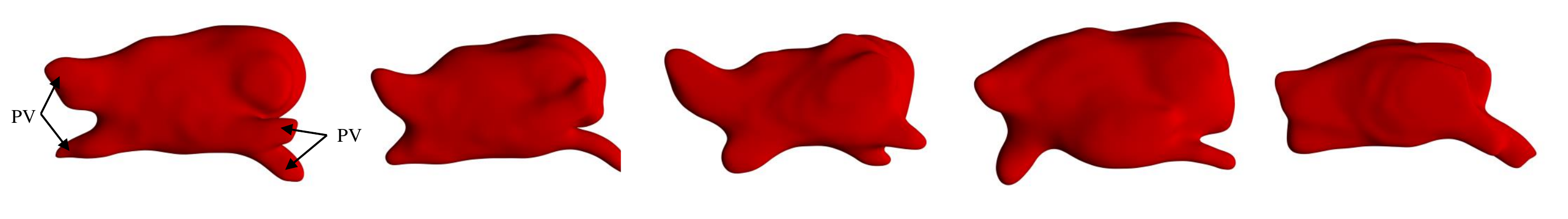}}
    \subfigure[] {\includegraphics[width=0.32\textwidth]{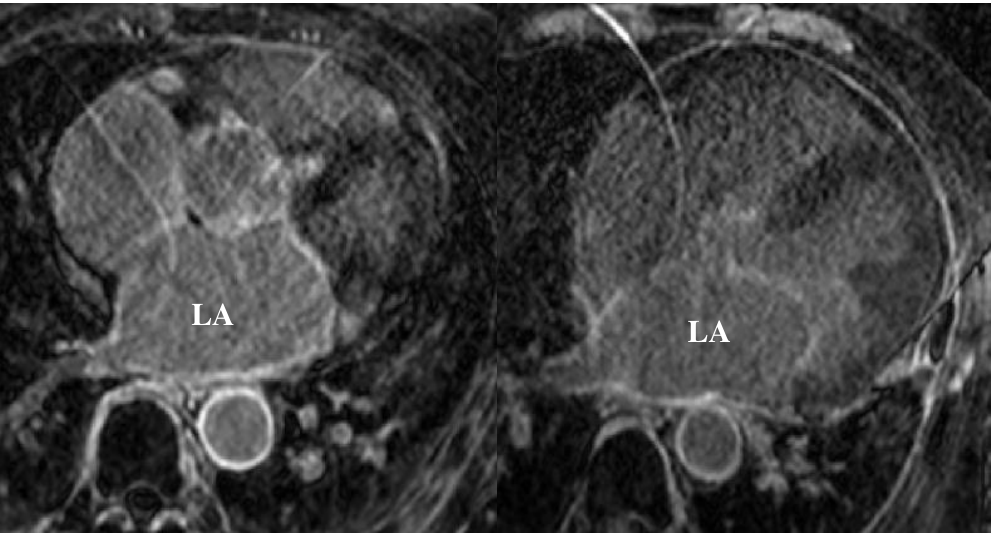}}
    \subfigure[] {\includegraphics[width=0.32\textwidth]{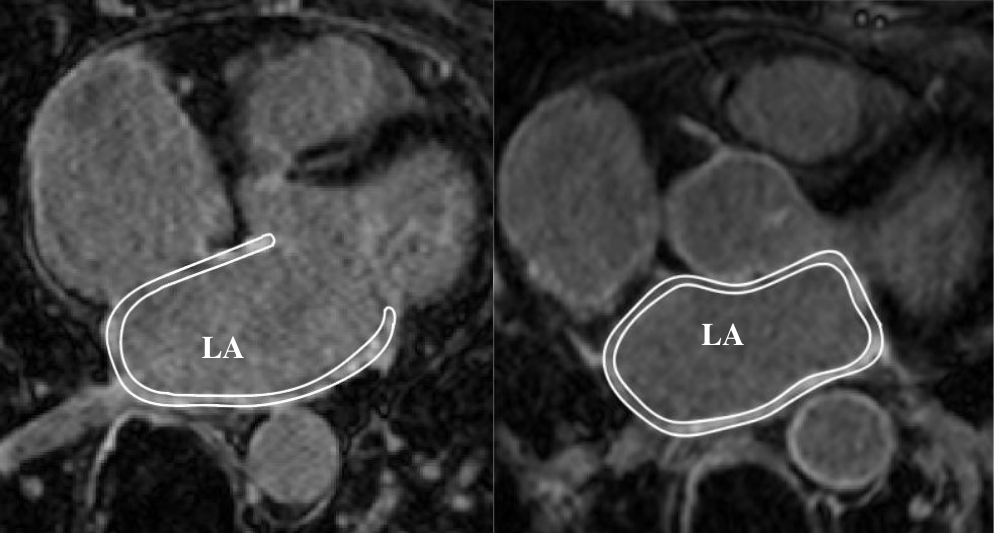}}
    \subfigure[] {\includegraphics[width=0.32\textwidth]{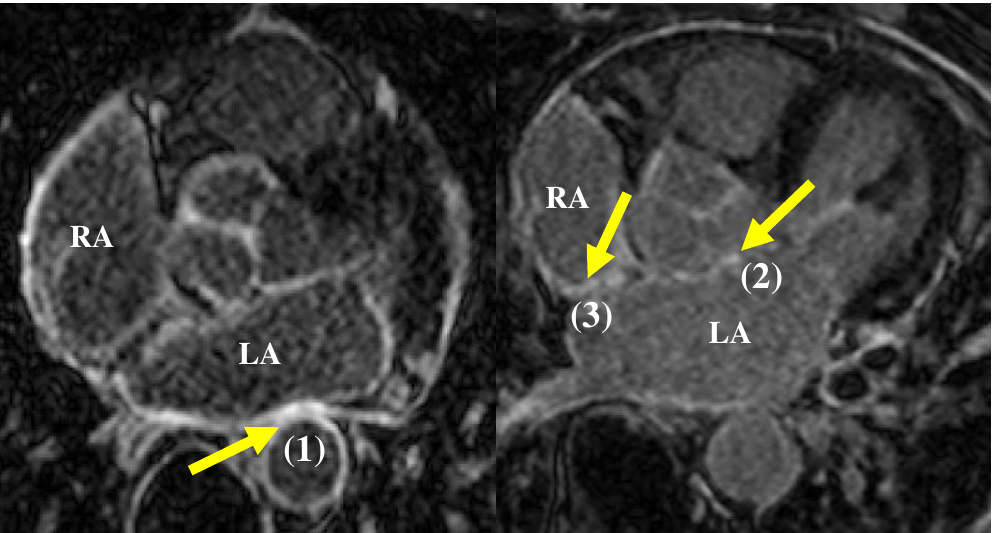}}
   \caption{The challenges of automatic segmentation and quantification of LGE MRI for AF:
     (a) various LA and pulmonary vein (PV) shapes;
     (b) two typical LGE MRIs with poor quality; 
     (c) thin atrial walls highlighted using bright white color in the figure;
     (d) surrounding enhanced regions pointed out by the arrows, where (1) and (2) indicate the enhanced walls of descending and ascending aorta, respectively; and (3) denotes the enhanced walls of the right atrium (RA).
     Images (b)-(d) adapted from \citet{journal/MedIA/li2020} \reviewerone{}{with permission}.}
\label{fig:intro:challenges}
\end{figure*}

Manual delineations of the LA, LA wall, scars, and ablation gaps are all labor-intensive and prone to be subjective, so their automation is highly desired, which nevertheless remains challenging.
The challenges for automatic LA cavity segmentation are mainly from the large variations in terms of LA shape, intensity range as well as poor image quality.
For the LA wall analysis, two additional difficulties are presented, i.e., the intrinsic thin wall thickness and the complex structure of the LA wall.
Here, the complex structure refers to the multiple openings in its 3D structure such as the pulmonary veins (PV) and mitral valve (MV) of the LA.
For the scar analysis, its unique challenge lies in the enhanced noise from surrounding tissues.
For the gap quantification, the large variability in PV morphology (position, orientation, size, thickness) and the robustness to scar segmentation changes are the two major concerns.
\Leireffig{fig:intro:challenges} illustrates and explains part of these challenges in an intuitive way.

\subsection{Study inclusion and literature search}
\begin{figure}[t]\center
    \includegraphics[width=0.45\textwidth]{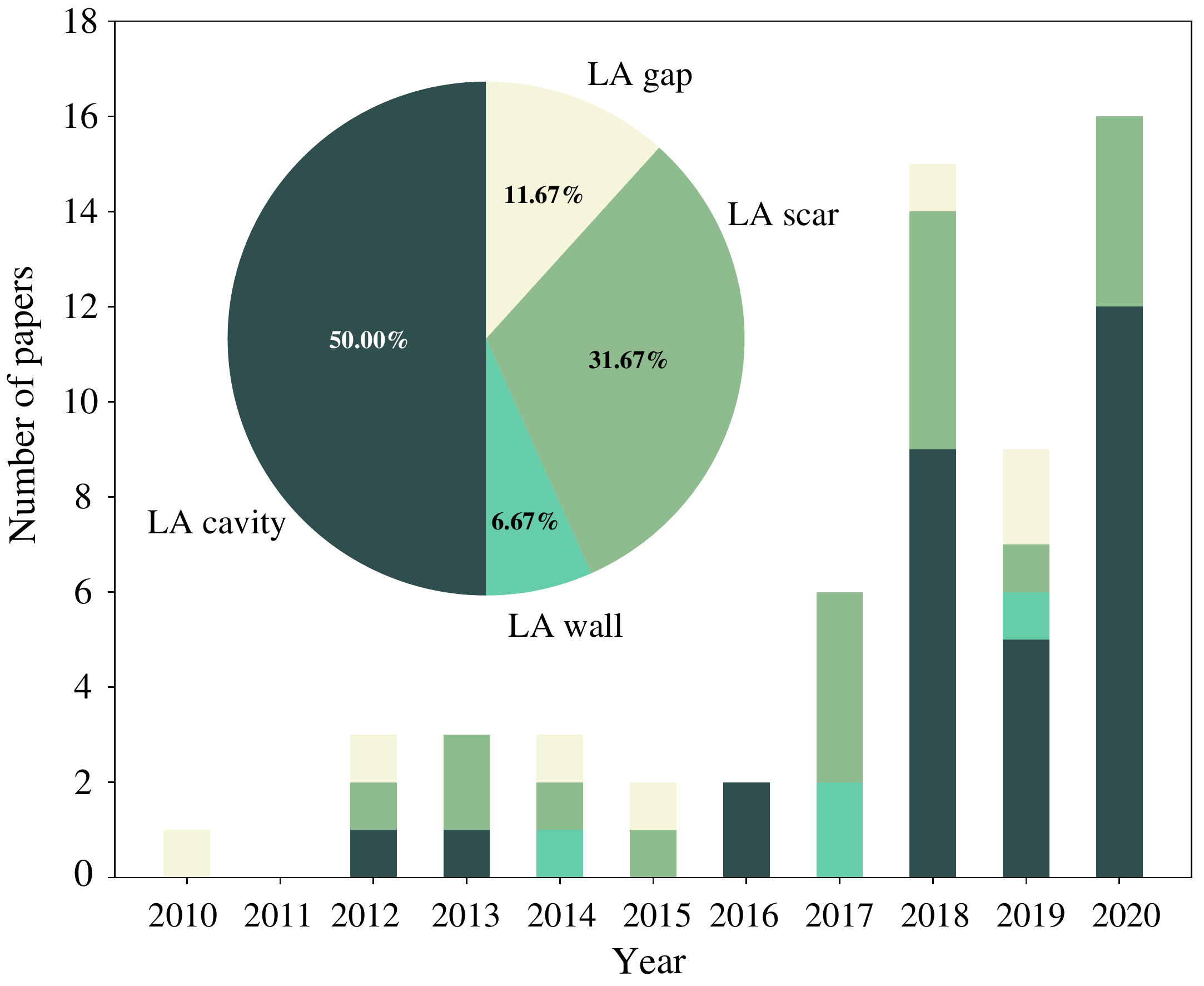}
   \caption{The distributions of papers of LGE MRI segmentation and quantification for AF patients per year and task.}
\label{fig:intro:review_distribution}
\end{figure}

In this work, we aim to provide the reader with a survey of the state-of-the-art image computing techniques, important results as well as the related literature for AF studies.
To ensure comprehensive coverage, we have screened publications from the last 10 years related to this topic.
Our main sources of references were Internet searches using engines such as Google Scholar, PubMed, IEEE-Xplore, and Citeseer.
To cover as many related works as possible, flexible search terms have been employed when using these search engines, as summarized in \Leireftb{tb:intro:search item}.
Both peer-reviewed journal papers and conference papers were included here.
We have also followed the references found in papers from these sites, and finally collected a comprehensive library of more than 130 papers. 
\Leireffig{fig:intro:review_distribution} presents the distributions of papers in segmentation and quantification from LGE MRI for AF patients per year/task.
Note that we generally picked the most detailed and representative ones for this review when we encountered several papers from the same authors about the same subject.

\begin{table} [t] \center
    \caption{Search engines and expressions used to identify potential papers for review. 
     }
\label{tb:intro:search item}
{\small
\begin{tabular}{p{0.75cm}|p{7.2cm}}
\hline
Engine                 & Google scholar, PubMed, IEEE-Xplore, and Citeseer\\
\hline
\multirow{7}{*}{Term}  &``Atrial fibrillation" or AF \textbf{and} \\ \cline{2-2} 
                              ~& ``Late gadolinium/ delayed enhancement/ contrast enhanced (cardiac) magnetic resonance" or ``LGE/ DE/ CE MR(I)" or ``LGE-/ DE-/ CE-MR(I)" or ``LGE/ DE/ CE CMR" or ``LGE-/ DE-/ CE-CMR"  \textbf{and}\\\cline{2-2} 
                              ~& Classif$^*$/ segment$^*$/ quantif$^*$/ localiz$^*$/ detect$^*$ \textbf{and}\\ \cline{2-2} 
                              ~&  ``Left atrium/ atrial" or LA  \textbf{or}\\\cline{2-2} 
                              ~&  ``Atrial wall/ myocardi$^*$" or ``wall thickness" \textbf{or}\\\cline{2-2} 
                              ~& ``Atrial scars/ fibrosis/ lesion" or ``ablation pattern" \textbf{or}\\\cline{2-2} 
                              ~&  ``Ablat$^*$"/ lesion gaps" or ``gaps in ablation lesion" or ``incomplete ablation pattern"\\
\hline
\end{tabular} }\\
\end{table}

\subsection{Related review literature}
\begin{figure}[t]\center
    \includegraphics[width=0.43\textwidth]{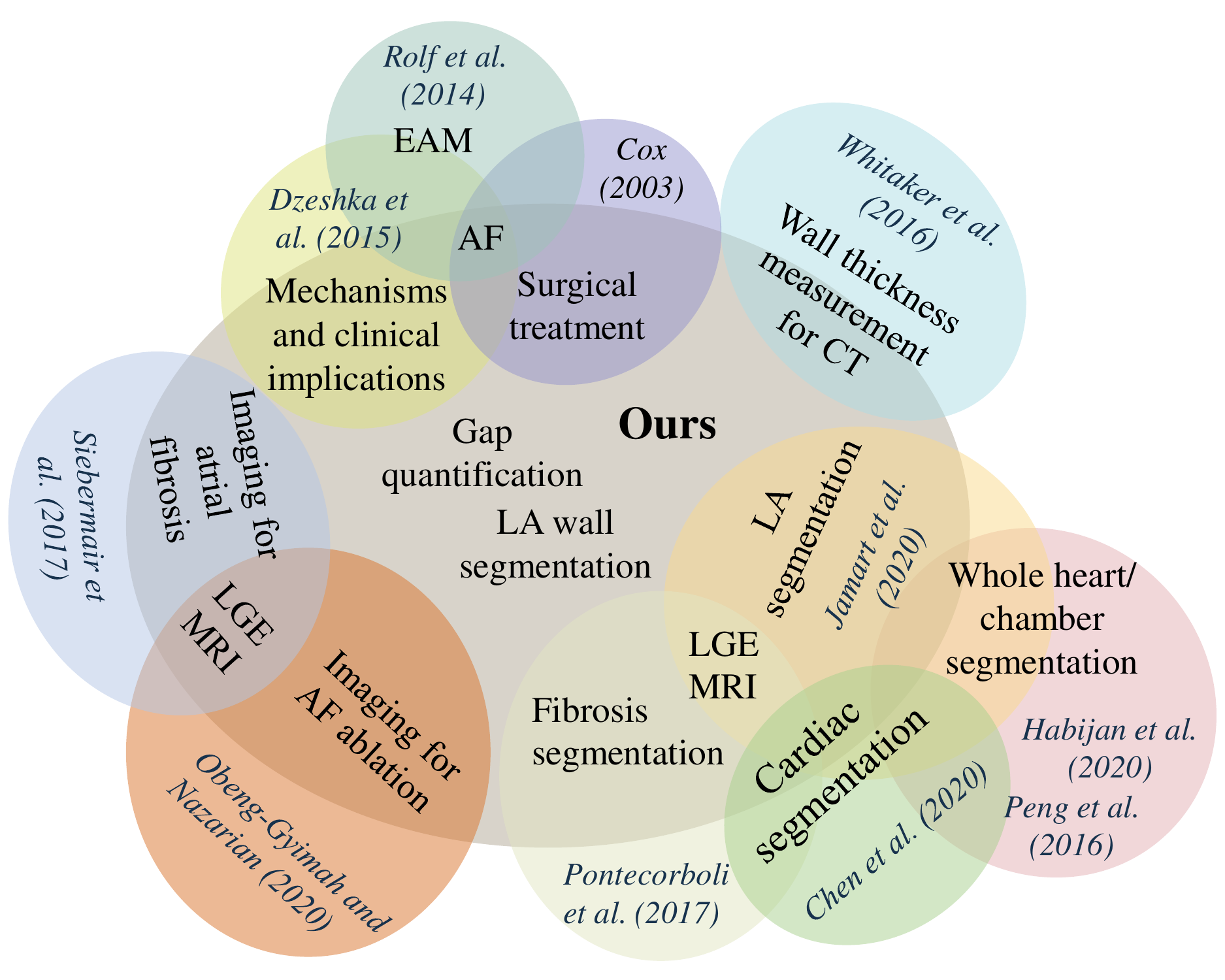}
   \caption{Comparison of the scopes of related review studies via Venn diagram.}
\label{fig:intro:review}\end{figure}

\Leireftb{tb:intro:review paper} lists existing review papers related to AF.
One can see that most current AF-related review papers focused on a clinical survey instead of the methodology of image computing, such as segmentation or quantification algorithms.
Only two reviews, \citet{journal/EP/pontecorboli2017} and \citet{journal/FCM/jamart2020}, are similar to ours in terms of the topic (LGE MRI) and style (technical).
However, only conventional thresholding methods or only deep learning (DL)-based methods were reviewed in each work.
\Leireffig{fig:intro:review} visualizes the scopes of current reviews as well as this review, and one can see that the scopes are different although partial overlaps can be found.
Besides, our review organizes the related works according to the clinical pipeline (see \Leireffig{fig:intro:review_structure}), resulting in an intuitive structure of the paper.

\begin{table*}\center
    \caption{Summary of the review papers related to AF. 
     EAM: electroanatomical mapping;
     JAF: Journal of Atrial Fibrillation;
     JACC: Journal of the American College of Cardiology;
     RMPBM: Magnetic Resonance Materials in Physics, Biology and Medicine;
     JICRM: The Journal of Innovations in Cardiac Rhythm Management;
     FCM: Frontiers in Cardiovascular Medicine;
     CET: Cardiovascular Engineering and Technology;
     DL: deep learning;
     CT: computed tomography.
     }
    \label{tb:intro:review paper}
    {\small
    \begin{tabular}{p{4.6cm}|p{1cm}p{5.6cm}p{5.2cm}}
    
    \hline
    Source	   & Venue &  Scope & Limitation \\
    \hline
    \citet{journal/EP/cox2003}	        & Europace & Surgical treatment of AF          & Clinical review \\
    \citet{journal/JAF/rolf2014}        & JAF      & EAM of AF                         & Clinical review \\
    \citet{journal/JACC/dzeshka2015}    & JACC     & Mechanisms and clinical implications of AF & Clinical review \\
    \citet{journal/EP/whitaker2016}	    & Europace & Wall thickness measurement for CT & Target image is not LGE MRI  \\
    \citet{journal/RMPBM/peng2016}      & RMPBM    & Cardiac chamber segmentation      & Target partially includes LA cavity \\
    \citet{journal/EP/pontecorboli2017} & Europace & Fibrosis segmentation from LGE MRI& Only thresholding methods are included \\
    \citet{journal/JACC/siebermair2017} & JACC     & LGE fibrosis imaging              & Clinical review \\
    \citet{journal/JICRM/obeng2020}	    & JICRM    & Imaging for AF ablation           & Clinical review \\
    \citet{journal/FCM/jamart2020}      & FCM      & LA cavity segmentation from LGE MRI      & Only DL-based methods are included \\
    \citet{journal/FCM/chen2020}        & FCM      & DL-based cardiac segmentation     & Target partially includes LA and its scars \\
    \citet{journal/CET/habijan2020}     & CET      & Whole heart and chamber segmentation & Target partially includes LA cavity \\
    \hline
\end{tabular}}\\
\end{table*}

\subsection{Structure of this review} \label{intro:review structure}
The remainder of the paper is organized as follows (compare \Leireffig{fig:intro:review_structure}):
Section \ref{AF imaging} presents the current common imaging tools used in AF ablation and the importance of LGE MRI in the management of AF.
Section \ref{method} systematically reviews the state-of-the-art image computing techniques and results of LA cavity, wall, scar, and ablation gap segmentation and quantification.
Section \ref{evaluation issue} presents the public data, evaluation measures, and state-of-the-art evaluation results on the public data for each task.
Potential clinical applications are provided in Section \ref{clinical aplication}.
Discussion of current LA LGE MRI computing challenges and future perspectives are given in Section \ref{discussion}, along with a conclusion in Section \ref{conclusion}.


\section{Imaging of atrial fibrillation} \label{AF imaging}
Medical images can offer crucial information for the evaluation and treatment of AF patients, and have been widely used in the ablation process \citep{journal/EHJ/tops2010,journal/JICRM/obeng2020}.
\Leireftb{tb:intro:imaging} summarizes the common imaging modalities used in three ablation stages (before, during, and after catheter ablation), mainly referring to \citet{journal/EHJ/tops2010} and \citet{journal/JICRM/obeng2020}.
One can see that diverse imaging modalities have been introduced in the ablation process, each of which assists in various aspects of the procedure.

\begin{table*} [t] \center
    \caption{The role of different imaging modalities in AF ablation procedures. 
    CA: catheter ablation; LAA: LA appendage; ICE: intracardiac echocardiography; TTE: transthoracic echocardiography; STE: speckle tracking echocardiography.
     }
\label{tb:intro:imaging}
{\small
\begin{tabular}{p{1.3cm}|p{3.6cm}p{2.3cm} p{9.4cm}}
\hline
Stage  & Target & Imaging modality & Important summary\\
\hline

\multirow{10}*{Before CA}&\multirow{2}*{Assessment of LAA thrombus} & TEE & Clinical reference for LAA thrombi identification \reviewerone{}{\citep{journal/Europace/calkins2007}} \\
~&~                                                                & CT/ MRI & Low inter-observer agreement \reviewerone{}{\citep{journal/AJR/mohrs2006,journal/JCE/gottlieb2008}} \\\cdashline{2-4}
~&\multirow{4}*{\tabincell{l}{Assessment of LA size and \\ anatomy}}   &  TTE        & The most commonly used imaging technique in daily clinical practice \reviewerone{}{\citep{journal/Heart/tops2007}} \\
~&~                                                  &  RT3DE/STE & New techniques for the assessment of LA volumes  \reviewerone{}{\citep{journal/CU/cameli2012}} \\
~&~                                                  &  MRI        & Gold standard for the assessment of LA volumes \reviewerone{}{\citep{journal/BRI/kuchynka2015}} \\\cdashline{2-4}
~                     & Assessment of PV anatomy     &  CT/ MRI    & Provides detailed 3D information on PV anatomy as a ``road-map" for ablation \reviewerone{}{\citep{journal/IJC/bhagirath2014}} \\\cdashline{2-4}
~&Assessment of fibrosis                             &  LGE MRI    & The most widely used MRI protocol for LA fibrosis imaging \reviewerone{}{\citep{journal/JACC/siebermair2017}} \\\hline
\multirow{10}*{During CA}& Positioning catheters      &  Fluoroscopy& Standard imaging modality in the electrophysiology laboratory; used to visualize catheters and devices \reviewerone{}{\citep{journal/Ep/bourier2016}} \\\cdashline{2-4}
~                       &Transseptal puncture        &  ICE        & Used to enhance the safety of transseptal puncture and catheter tissue contact; used to visualize inter-atrial septum and puncture needle \reviewerone{}{\citep{journal/Heart/jongbloed2005}} \\\cdashline{2-4}
~&\multirow{5}*{Visualization of LA and PVs}         & Fluoroscopy & New rotational angiography technique to accurately identify PV anatomy and diameters \reviewerone{}{\citep{journal/IJC/thiagalingam2008}} \\
~&~                                                  & ICE         & Real-time assessment of PV ostium with a limitation on the detection of small proximal branches from PVs \reviewerone{}{\citep{journal/JCE/saad2002,journal/AJC/wood2004,journal/JACC/jongbloed2005}}\\\hline
\multirow{16}*{After CA}& Assessment of PV stenosis  &  CT/ MRI    & Preferably, these 3D techniques are correlated with pre-procedural images for detection of PV stenosis \reviewerone{}{\citep{journal/JACC/holmes2009}} \\\cdashline{2-4}
~                       &Detection of pericardial    &  TTE        & Routine echocardiography should be performed before discharge and during the follow-up study \reviewerone{}{\citep{journal/Europace/calkins2007}} \\\cdashline{2-4}

~                       &Esophageal injury           & CT/ MRI    & Performed when atrio-oesophageal fistula is suspected \reviewerone{}{\citep{journal/Europace/calkins2007}} \\\cdashline{2-4}

~&\multirow{5}*{\tabincell{l}{Assessment of LA size \\ and function}}  & TTE         & Conventional method for the detection of LA volumes and function \reviewerone{}{\citep{journal/AJC/blondheim2005}}\\ %
~&~                                                  & RT3DE/CT/(LGE) MRI & 3D assessment of LA volumes allows the detection of LA reverse remodelling \reviewerone{}{\citep{journal/BMC/zhang2017,journal/SRA/polaczek2019,journal/JCE/tsao2005,journal/CAE/mcgann2014}} \\\cdashline{2-4}
~&Assessment of wall thickness                       &  TEE/CT/(LGE) MRI & Increased atrial wall thickening was seen in the post-ablation scans \reviewerone{}{\citep{journal/IJC/nakamura2011,journal/MedIA/karim2018,journal/CCI/habibi2015}} \\\cdashline{2-4}
~&\multirow{3}*{Assessment of scars and gaps}        & LGE MRI     & Promising in the ablation lesion visualization \reviewerone{}{\citep{journal/JACC/mcgann2008}}\\
~&~                                                  & T1 mapping MRI & New technique without contrast agent for the assessment of scars \reviewerone{}{\citep{journal/HR/beinart2013}}\\
\hline
\end{tabular} }\\
\end{table*}

\subsection{Imaging for ablation procedures}
Before catheter ablation (CA), the first step is to exclude contraindication, such as the LA appendage (LAA) thrombi which are normally detected using transesophageal echocardiography (TEE) \citep{journal/AJC/ellis2006,journal/Europace/calkins2007,journal/JACC/pathan2018}.
\reviewerone{}{MRI and computed tomography (CT) can be used to detect LA thrombi, but both tend to have a low inter-observer agreement \citep{journal/AJR/mohrs2006,journal/JCE/gottlieb2008}.
In addition, the images are statically acquired a few seconds after the arrival of contrast to the LAA. Hence, it could be difficult to differentiate LAA thrombi from sluggish flow \citep{journal/Circulation/romero2013}.}
To select patients expected for successful CA, the assessment of LA, PVs, and fibrosis are the key steps \citep{journal/EHJ/berruezo2007,journal/JCE/akoum2011}.
Three-dimensional (3D) imaging techniques, such as CT and MRI, are generally used for PV anatomy assessment.
PV anatomy can also be measured by TEE, achieving up to 95\% concordance with MRI \citep{journal/JCM/toffanin2006}.
Moreover, cardiac MRI remains the gold standard for fibrosis assessment \citep{journal/JICRM/obeng2020}.
Especially, LGE MRI appears to be a promising alternative for pre-ablation scar visualization and quantification \citep{journal/JACC/siebermair2017}.

During CA, fluoroscopy is the most commonly employed imaging technique in the electrophysiology laboratory.
Intracardiac echocardiography (ICE) offers real-time imaging of the PVs and adjacent structures and enhances the safety of transseptal puncture by visualizing inter-atrial septum and puncture needle \citep{journal/Heart/jongbloed2005}.
Both ICE and fluoroscopy can visualize the LA and PVs \citep{journal/JCE/saad2002}.
Note that the integration of different imaging modalities during CA is promising \citep{journal/EHJ/tops2010}, but is out of the scope of this review. 

After CA and during the follow-up study, the main target of post-procedural imaging is to monitor complications and help predict recurrence.
The most frequently occurring complications of AF ablation include PV stenosis, pericardial effusion, and atrio-oesophageal fistul.
Multi-slice CT and MRI are usually used for accurate assessment of PV stenosis and esophageal injury \citep{journal/JACC/holmes2009}.
Transthoracic echocardiography (TTE) is a recommended imaging tool for screening to detect pericardial effusion \citep{journal/Europace/calkins2007}.
To predict recurrence, LA size and functions are important indices, as LA ablation can lead to the formation of scars and subsequent changes in LA anatomy \citep{journal/JACC/casaclang2008}.
\textit{For the follow-up analysis of LA volumes}, TTE is typically used, but 3D techniques, such as real-time 3D echocardiography \citep{journal/BMC/zhang2017}, multi-slice CT \citep{journal/SRA/polaczek2019}, and MRI \citep{journal/JCE/tsao2005}, especially LGE MRI \citep{journal/CAE/mcgann2014}, may provide more accurate information.
\textit{For the measurement of LA wall thickness}, \reviewerone{}{TEE has the advantages of high temporal resolution and short acquisition time, but it is difficult to obtain descriptive information on the LA wall due to its low spatial resolution \citep{journal/IJC/nakamura2011}.}
CT is an ideal modality, thanks to its high resolution, and MRI is widely considered to be the gold standard for the viability assessment of wall pathology \citep{journal/MedIA/karim2018}.

LGE MRI has been recently widely explored for scar and ablation gap quantification \citep{journal/MedIA/nunez2019,journal/JA/mishima2019}.
\reviewerone{}{Note that T1 mapping MRI could be used to obtain valuable imaging-based biomarkers for diffused cardiac fibrosis, which has been validated against histological studies \citep{journal/Radiology/sibley2012}.
For example, it is possible with T1 mapping to non-invasively quantify myocardium extracellular volume fraction, which is a biomarker of diffuse reactive fibrosis \citep{journal/JACC/taylor2016}.}
Nevertheless, it can be difficult to localize fibrosis using T1 mapping MRI, and it is therefore not appropriate for ablation procedure guidance or ablation gap identification. 
LGE MRI remains a promising method to detect focal and cohesive fibrosis \citep{journal/EP/pontecorboli2017}.

\begin{table*} [t] \center
    \caption{Imaging parameters for the LGE scar assessment utilized in several leading centers worldwide. 
    SA: Siemens Avanto;
    PA: Philips Achieva;
    TR: repetition time;
    TE: echo time;
    Acq. T: acquired time after contrast agent injection;
    CARMA: Comprehensive Arrhythmia Research and Management;
    DECAAF: Delayed-Enhancement MRI Determinant of Successful Radiofrequency Catheter Ablation of Atrial Fibrillation.
    Here, $^\dagger$ refers to multiple centers.
     }
\label{tb:imaging:scan param}
{\small
\begin{tabular}{l| llllll}
\hline
Source    & Center & Scanner & TR/TE (ms) & Acq. T (min) & Gadolinium dose &  Spacing (mm$^3$)\\
\hline
\citet{journal/CAE/badger2010}  & Utah, USA & 1.5 T, SA & 5.5/2.3   & 15     & 0.1 mmol/kg & 1.25 $\times$ 1.25 $\times$ 2.5  \\
\citet{journal/HR/taclas2010}   & Boston, USA & 1.5 T, PA & 3.8/1.52 & 15$\sim$25  & 0.2 mmol/kg & 1.3 $\times$ 1.3 $\times$ 4.0/5.0  \\
\citet{journal/JCE/hunter2013}  & Imperial/Barts, UK & 1.5 T, PA  & N/A & 20   & 0.4 mmol/kg & 1.5 $\times$ 1.5 $\times$ 4.0  \\
\citet{journal/JACC/bisbal2014} & Barcelona, Spain & 3 T  & 2.3/1.4  & 25$\sim$30  & 0.2 mmol/kg & 1.25 $\times$ 1.25 $\times$ 2.5  \\ 
\citet{journal/CAE/mcgann2014}  & CARMA$^\dagger$ & 1.5 T; 3 T, SA & 5.2/2.4; 3.1/1.4 & 5$\sim$9; 6$\sim$12  & 0.1 mmol/kg & 1.25 $\times$ 1.25 $\times$ 2.5 \\
\citet{journal/HR/fukumoto2015} & John Hopkins, USA  & 1.5 T, SA & 3.8/1.52  & 10$\sim$32  & 0.2 mmol/kg & 1.3 $\times$ 1.3 $\times$ 2.0  \\
\citet{journal/CAE/harrison2015}& KCL, UK & 1.5 T, PA & 6.2/3.0  & 20  & 0.2 mmol/kg & 1.3 $\times$ 1.3 $\times$ 4.0  \\
\citet{journal/JCE/akoum2015}   & DECAAF$^\dagger$ & 1.5 T; 3 T & 5.2/2.4; 3.1/1.4 & 15  & 0.1$\sim$0.2 mmol/kg & 1.25 $\times$ 1.25 $\times$ 2.5 \\
\citet{journal/JCE/cochet2015}  & Bordeaux, France & 1.5 T, SA & 6.1/2.4 & 15$\sim$30  & 0.2 mmol/kg & 1.25 $\times$ 1.25 $\times$ 2.5 \\
\hline
\end{tabular} }\\
\end{table*}

\subsection{LGE MRI for AF studies} \label{LGE MRI}
LGE MRI is mainly used to evaluate fibrosis and scars of AF patients before and after ablation.
This is because LGE MRI can discriminate scarring and healthy tissues by their altered wash-in and wash-out contrast agent kinetics \citep{journal/JAMA/marrouche2014}.
Scars are thus visualized as the regions of being enhanced or high signal intensity compared to healthy tissues \citep{journal/MP/yang2018}.
There is still no consensus on the option and dosage of the contrast agent, nor on the timing of image acquisition after contrast administration, as \Leireftb{tb:imaging:scan param} shows.  
Among the listed protocols, the DECAAF (Delayed-Enhancement MRI Determinant of Successful Radiofrequency Catheter Ablation of Atrial Fibrillation) protocol can be considered the most widely used one for LA fibrosis imaging \citep{journal/JACC/siebermair2017}.
Considering the importance and advances of LGE MRI in AF studies, in this review we mainly focus on the computing works on LGE MRI.

\section{Image computing} \label{method}

\begin{figure*}[t]\center
    \includegraphics[width=0.98\textwidth]{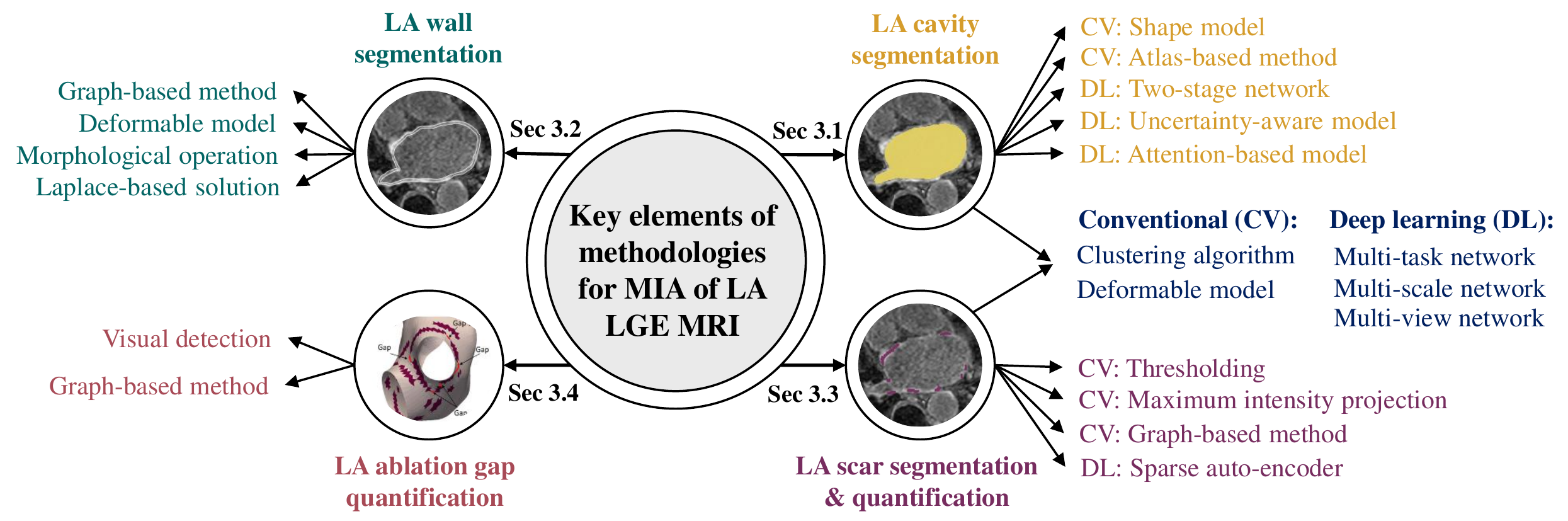}
   \caption{\reviewerone{}{Key elements of LA LGE MRI computing methodologies on the four tasks. MIA: medical image analysis.} }
\label{fig:method:summary}
\end{figure*}

We structure the review of image computing methodology according to the segmentation and quantification tasks in question, as presented in \Leireffig{fig:intro:review_structure}.
\reviewerone{}{To understand the key elements of  methodologies, we further classify the methods applied in each task (see \Leireffig{fig:method:summary}).
In the following sections, we will elaborate and discuss these methods and the corresponding results of different tasks in detail.
}

\subsection{LA cavity segmentation} \label{method: LA cavity seg}

\begin{table*} [t] \center
    \caption{Summary of previously published works on the automatic \textit{LA cavity segmentation} from LGE MRI.
     CNN: convolutional neural network; MAS: multi-atlas segmentation; MA-WHS: multi-atlas whole heart segmentation; 
     SVM: support vector machine; KNN: K nearest neighbor; 
     FCN: fully convolutional network; HAANet: hierarchical aggregation network; 
     ASD: average surface distance; 95HD: $95\%$ Hausdorff distance; VO: volume overlap; Jc: Jaccard; 
     Acc: Accuracy; Sen: Sensitivity; Spe: Specificity; Cf: Conform; APD: average perpendicular distance; Diam. Err: antero-posterior diameter error; Volume Err: volume error.
     Note that the results in studies labeled via $^{\ddagger}$ are from the benchmark paper \citep{journal/MedIA/xiong2020} for a fair comparison.
     The reported values in their manuscript may be inconsistent with the results reported in \citet{journal/MedIA/xiong2020} as they may employ parts of training data from the challenge as test data for evaluation.
     }
\label{tb:method:LA}
{\small
\begin{tabular}{p{3.4cm}|p{0.4cm}p{1.3cm}p{5.6cm}p{3cm}p{1.8cm}}
\hline
Study & Num & Pre/ Post & Algorithm & Evaluation & Dice$_\text{LA}$\\
\hline
\citet{conf/MI/gao2010}           & 20  & Post + Pre & Active contours + statistical shape learning             & N/A              &N/A\\  
\citet{conf/MICCAI/kutra2012}     & 59  & Pre        & Multi-model based fitting + SVM          & Acc  &N/A \\
\citet{journal/TIP/zhu2013}       & 64  & Post + Pre & Variational region growing + shape prior & Dice, VO, 95HD, ASD      &$0.79  \pm 0.05$\\  
\citet{conf/SMC/deng2016}         & 64  & Post + Pre & KNN + super pixel voting                 & Dice, VO                 &$0.81  \pm 0.08$\\  
\citet{journal/JMRI/tao2016}      & 56  & Pre        & MAS + 3D level-set                       & Dice, ASD                &$0.86  \pm 0.05$\\  
\citet{conf/STACOM/nunez2018}$^{\ddagger}$     & 154 & Post + Pre & MA-WHS + shape clustering                & Dice, HD, ASD            &$0.859 \pm 0.061$\\ 
\citet{conf/STACOM/qiao2018}$^{\ddagger}$      & 154 & Post + Pre & MAS + level-set                          & Dice, APD                &$0.861 \pm 0.036$\\  
\citet{journal/MedIA/li2020}      & 58  & Post       & MA-WHS                                   & Dice                     &$0.898 \pm 0.044$\\ 
\hline \hline
\citet{conf/MICCAI/chen2018}      & 100 & Post + Pre & Multi-view two-task network              & Dice, Acc, Spe, Sen      &$0.908 \pm 0.031$\\
\citet{journal/TMI/xiong2018}     & 154 & Post + Pre & Dual CNNs                                & Dice, HD, Spe, Sen       &$0.942 \pm 0.014$\\
\citet{conf/STACOM/chen2018}$^{\ddagger}$      & 154 & Post + Pre & Multi-task 2D U-Net          & Dice, Jc, HD, ASD        &$0.921 \pm 0.026$\\
\citet{conf/STACOM/vesal2018}$^{\ddagger}$     & 154 & Post + Pre & 3D U-Net+ dilated + residual & Dice, Jc, Acc            &$0.925 \pm 0.027$\\
\citet{conf/STACOM/savioli2018}$^{\ddagger}$   & 154 & Post + Pre & 3D volumetric FCN            & Dice, HD                 &$0.851 \pm 0.051$\\
\citet{conf/STACOM/li2018}$^{\ddagger}$        & 154 & Post + Pre & Attention based 3D HAANet    & Dice                     &$0.923 \pm 0.029$\\
\citet{conf/STACOM/bian2018}$^{\ddagger}$      & 154 & Post + Pre & ResNet101 + 2D pyramid Network    & Dice, Cf, Jc, HD, ASD    &$0.926 \pm 0.022$\\
\citet{conf/STACOM/puybareau2018}$^{\ddagger}$ & 154 & Post + Pre & VGG-16 + transfer learning + ``pseudo-3D”& Dice         &$0.923 \pm 0.023$\\ 
\citet{conf/STACOM/liu2018}$^{\ddagger}$       & 154 & Post + Pre & 2D U-Net + FCN               & Dice                     &$0.903 \pm 0.032$\\
\citet{conf/STACOM/preetha2018}$^{\ddagger}$   & 154 & Post + Pre & 2D U-Net                     & Dice                     &$0.887 \pm 0.031$\\
\citet{conf/STACOM/de2018}$^{\ddagger}$        & 154 & Post + Pre & 2D U-Net                     & Dice                     &$0.897 \pm 0.035$\\
\citet{conf/STACOM/jia2018}$^{\ddagger}$       & 154 & Post + Pre & Two-stage 3D U-Net + contour loss & Dice, HD, Spe, Sen  &$0.907 \pm 0.031$\\
\citet{conf/STACOM/xia2018}$^{\ddagger}$       & 154 & Post + Pre & Two-stage 3D V-net           & Dice                     &$0.932 \pm 0.022$\\
\citet{conf/STACOM/yang2018}$^{\ddagger}$      & 154 & Post + Pre & Two-stage 3D U-Net + transfer learning   & Dice, Cf, Jc, HD, ASD    &$0.925 \pm 0.023$\\
\citet{conf/STACOM/borra2018}     & 154 & Post + Pre & Otsu’s algorithm + 3D U-Net              & Dice                     &$0.898$\\
\citet{conf/STACOM/jamart2019}    & 154 & Post + Pre & Two-stage 2D V-net                       & Dice, Jc, HD, ASD, Diam. Err, Volume Err &$0.937$\\
\citet{conf/MICCAI/yu2019}        & 100 & Post + Pre & Uncertainty-aware model                  & Dice, Jc, 95HD, ASD      &$0.889$\\
\citet{conf/MI/wang2019}          & 100 & Post + Pre & Ensembled U-Net                          & Dice                     &$0.921 \pm 0.020$\\
\citet{journal/JCMARS/du2020}     & 100 & Post + Pre & Multi-scale dual-path network            & Dice, Cf, Jc, HD         &$0.936 \pm 0.005$\\
\citet{journal/QIMS/borra2020}    & 100 & Post + Pre & 2D/ 3D U-Net                             & Dice, HD, Spe, Sen       &$0.895 \pm 0.025$/ $0.914 \pm 0.015$\\
\citet{conf/AIPR/xiao2020}        & 100 & Post + Pre & Multi-view network                       & Dice, HD, ASD            &$0.912$\\
\citet{conf/ICPR/zhao2021}        & 100 & Post + Pre & ResNet101 + hybrid loss                  & Dice, 95HD, ASD          &$0.918 \pm 0.011$\\
\citet{journal/MedIA/li2021}      & 60  & Post       & Multi-task 3D U-Net + spatial encoding   & Dice, HD, ASD            &$0.913 \pm 0.032$\\
\hline
\end{tabular} }\\
\end{table*}

\reviewerone{}{In recent years, many algorithms have been proposed to perform automatic LA cavity segmentation from medical images, but mostly for non-enhanced imaging modalities.
Conversely, a limited number of works for the LA cavity segmentation from LGE MRI were reported in the literature before 2018.
Most of the current studies on the LA cavity segmentation from LGE MRI are still based on time-consuming and error-prone manual segmentation methods \citep{journal/JACC/higuchi2018,journal/EP/njoku2018}.
This is mainly because LA cavity segmentation methods in non-enhanced imaging modalities are difficult to directly apply to LGE MRI, due to the existence of contrast agents and low-contrasted boundaries.
Existing conventional automatic LA LGE MRI segmentation approaches generally require additional information, such as shape priors \citep{journal/TIP/zhu2013} or \reviewerone{}{other images, such as non-enhanced 3D MRI \citep{journal/MedIA/li2020} and contrast enhanced magnetic resonance angiogram (MRA) \citep{journal/TMI/ravanelli2014,journal/JMRI/tao2016,journal/Frontiers/roney2020}}.
Recently, with the development of DL in medical image processing, numerous DL-based algorithms are proposed for the automatic LA cavity segmentation directly from LGE MRI \citep{journal/MedIA/xiong2020}.
\Leireftb{tb:method:LA} summarizes the representative methods and their results in chronological order.
The upper and lower parts of the table summarize conventional (non-DL-based methods) and DL-based methods, respectively.}

\subsubsection{\reviewerone{}{Conventional methods for LA cavity segmentation}}
\reviewerone{}{Conventional methods for LA cavity segmentation can be classified into four kinds, i.e., shape models, clustering algorithms, deformable models (region growing, activate contour, and level-set), and atlas-based methods.}

\begin{figure}[t]\center
    \includegraphics[width=0.48\textwidth]{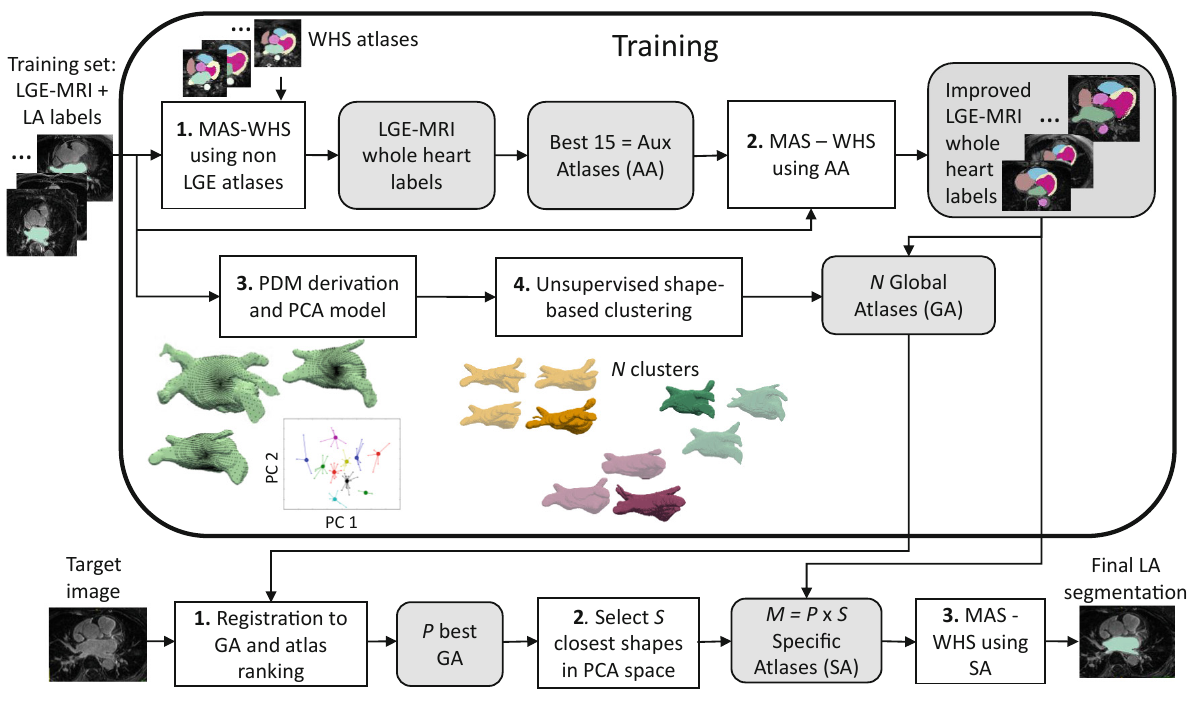}
   \caption{\reviewerone{}{A framework example for the LA cavity segmentation from \citet{conf/STACOM/nunez2018}. 
   The LA cavity segmentation was achieved via multi-atlas whole heart segmentation (MAS-WHS) and shape modeling of the LA.
   Image adapted from \citet{conf/STACOM/nunez2018} with permission.} }
\label{fig:method example:LA cavity}
\end{figure}

\paragraph{Shape models/ clustering algorithms}
\reviewerone{}{Many works incorporated anatomical or shape priors to improve the robustness against the large variability of LA shapes and intensity distributions.
For example, \citet{conf/MI/gao2010} used shape learning and region-based active contour evolution for the LA cavity segmentation.
The shape learning aimed to utilize prior shape knowledge, to solve the unclear boundary problem in LGE MRI when using the active contour method.
\citet{journal/TIP/zhu2013} achieved the LA cavity segmentation using a variational region growing with a moments-based shape prior.
They adjusted the weights between the data-driven term and shape prior constraint to adapt for the changes in the volume of the target region.
\citet{conf/STACOM/nunez2018} constructed LGE MRI atlases via multi-atlas segmentation (MAS) and then clustered the LA shapes using principal component analysis to perform a second MAS for the LA cavity segmentation, as presented in \Leireffig{fig:method example:LA cavity}.
It remains too complicated so far to cover the large shape variation between LA cavities of different subjects by simply imposing a shape prior.}

\paragraph{Deformable models}
\reviewerone{}{The major challenge of deformable models on the LA cavity segmentation arises from the wide variability of the intensity distribution in LGE MRI.
To solve this, \citet{journal/TIP/zhu2013} designed a variational region growing method to reduce its sensitivity to the change of intensity distribution.
The seed search in their work was performed by incorporating certain geometric information of PVs relative to the LA.
Instead of performing global optimization, \citet{journal/JMRI/tao2016} and \citet{conf/STACOM/qiao2018} employed level-set for local refinement on the global segmentation obtained by MAS. 
The advantage of deformable models is that they do not have a prior assumption about the object geometry and are therefore skillful at capturing local shape variations, such as the PV regions of the LA.
Therefore, it is effective to combine deformable models for local attention with other models considering the global shape information of LA.
Examples include \citet{conf/MI/gao2010} and \citet{journal/TIP/zhu2013} where a shape prior was employed as a global constraint.}

\paragraph{Atlas-based methods}
\reviewerone{}{An alternative way is to use atlas-based methods that can be robust to the LA cavity with high anatomical variations. 
For instance, \citet{journal/JMRI/tao2016} and \citet{journal/MedIA/li2020} utilized atlas-based methods employing the label of another image (from the same patient) with better anatomical information to assist the LA cavity segmentation of LGE MRI.
\citet{journal/JMRI/tao2016} employed MAS to segment the LA cavity from the MRA, and then mapped the generated label to LGE MRI followed by a level-set based refinement.
They compared the results with that of solely using LGE MRI (directly employing MAS on LGE MRI) and found that the former achieved better results.
They also tested their method on the public dataset from the \textit{Atrial Segmentation Challenge} where only LGE MRI was provided \citep{conf/STACOM/qiao2018}, and achieved better performance in terms of Dice compared to that in \citet{journal/JMRI/tao2016} ($0.88 \pm 0.03$ vs. $0.86 \pm 0.05$).
This may be due to the difference in the dataset, as the public data includes both pre- and post-ablation images.
Similarly, \citet{journal/MedIA/li2020} employed an auxiliary MRI sequence to assist the LA cavity segmentation of LGE MRI using MAS methods and obtained a better Dice score ($0.898 \pm 0.044$) than other conventional methods.
Particularly, \citet{journal/MedIA/li2020} and \citet{conf/STACOM/nunez2018} adopted a multi-atlas based whole heart segmentation (MA-WHS) and then extracted the LA sub-structure.
This is because the LGE MRIs employed in their studies cover the whole heart, and MA-WHS could be helpful to exclude surrounding sub-structures of LA.
Although in clinical routine LGE MRI may have limited field-of-view, all current public LA LGE MRI datasets were specifically acquired to cover the whole heart with the development of novel whole-heart high-resolution LGE techniques \citep{journal/JMRI/toupin2021}.
Although auxiliary images can provide better anatomical information, the anatomy extracted from them may be highly deformed compared to that acquired from LGE MRI. 
It may cause difficulties in the co-registration step and lead to subsequent incorrect segmentation of the LA cavity.
Moreover, conventional atlas-based methods are generally time-consuming due to multiple image registration steps.}


\subsubsection{\reviewerone{}{Deep learning-based methods for LA cavity segmentation}}
\reviewerone{}{For the LA cavity segmentation, many basic neural network architectures have been employed.
To boost the feature learning ability of networks, a series of works have focused on optimizing network structures, investigating different loss functions, and applying anatomical constraints.
Here, we mainly classify these DL-based methods according to the network architectures, and will also discuss the loss functions and anatomical constraints used to train the networks.}

\paragraph{Architecture of network}
\reviewerone{}{Recently, many methods based on different network structures were developed with the launch of the \textit{Atrial Segmentation Challenge} in MICCAI 2018, where U-Net was commonly employed as the backbone.
For example, \citet{conf/STACOM/vesal2018} employed a 3D U-Net with dilated convolutions at the bottom of the network and residual connections between encoder blocks, to incorporate both local and global knowledge.
\citet{conf/STACOM/li2018} proposed an attention-based hierarchical aggregation network for the LA cavity segmentation, and the basic network is a 3D U-Net.
\citet{journal/QIMS/borra2020} tested both 2D and 3D U-Net for the LA cavity segmentation and found that 3D pipelines showed significantly better performance compared to the 2D pipelines.
\citet{conf/MI/wang2019} utilized ensemble attention U-Net, dense U-Net, and residual U-Net models to segment LA. 
\citet{conf/STACOM/liu2018}, \citet{conf/STACOM/preetha2018}, and \citet{conf/STACOM/de2018} all employed 2D U-Net for the LA cavity segmentation, and \citet{conf/STACOM/liu2018} also tested the performance of fully convolutional networks (FCNs).
Instead of using U-Net as the backbone, \citet{conf/STACOM/bian2018} used ResNet101 for the LA cavity segmentation and adopted a pyramid module to learn multi-scale semantic information in the feature map.
\citet{conf/STACOM/puybareau2018} achieved the LA cavity segmentation by transfer learning from VGG-16, a pre-trained network used to classify natural images. 
\citet{conf/STACOM/savioli2018} presented a 3D volumetric FCN for the LA cavity segmentation. 
Besides the architecture, \citet{journal/FCM/jamart2020} emphasized the importance of relevant loss function selection for the LA cavity segmentation.
\citet{conf/STACOM/jia2018} proposed a novel contour loss function to include distance information for good shape consistency.
\citet{conf/ICPR/zhao2021} employed a hybrid loss to focus on the boundaries as much as on regions, and therefore reduced the impact of noisy neighboring tissues.
\citet{journal/MedIA/li2021} introduced a spatial encoding (SE) loss to incorporate continuous spatial information of the LA.
Their experiments showed that the SE loss could be effective to remove noisy patches in the final predicted segmentation, and therefore evidently reduced the Hausdorff distance (HD) value.
For the loss function selection, one could refer to the review paper \citep{journal/MedIA/ma2021}, where Dice-related compound loss functions were recommended for medical image segmentation tasks.}

\paragraph{Multi-task networks}
\reviewerone{}{Multi-task learning has been adopted for the LA cavity segmentation to utilize its possible relationship with other auxiliary tasks. For example, \citet{conf/MICCAI/chen2018} and \citet{journal/MedIA/li2021} performed simultaneous LA cavity and scar segmentation via multi-task learning.
The simultaneous optimization scheme showed better performance than solving the two tasks independently which ignored the intrinsic spatial relationship between the LA cavity and scars. 
\citet{conf/STACOM/chen2018} designed a two-task network for both LA cavity segmentation and pre/ post ablation image classification to learn additional anatomical information.
The results indicated that multi-task learning obtained better segmentation performance compared to baseline U-Net method training with a single segmentation task.}

\paragraph{Two-stage networks}
\reviewerone{}{A two-stage training strategy has been gradually employed to replace conventional pre-processing (such as the Otsu’s algorithm employed in \citet{conf/STACOM/borra2018}) for the region of interest (ROI) extraction.
For instance, \citet{conf/STACOM/jia2018}, \citet{conf/STACOM/xia2018}, \citet{conf/STACOM/yang2018}, and \citet{conf/STACOM/jamart2019} all utilized two-stage U-Net/ V-Net and achieved top performances in the LA cavity segmentation.
The first stage was to roughly locate the LA cavity center for ROI extraction, while the second stage was to perform the LA cavity segmentation from the cropped ROI.
In this way, a memory-efficient and accurate framework was developed, and the class imbalance problem was also mitigated.
It is worth mentioning that \citet{conf/STACOM/xia2018} obtained the first-ranked results (mean Dice score of $0.932 \pm 0.022$) in \textit{Left Atrium Segmentation Challenge} by using the two-stage network.}

\paragraph{Multi-view networks} \label{LA method: multi-view network}
\reviewerone{}{The major drawback of 2D networks is that they ignore the inter-slice correlation in the 3D LGE MRI.
To solve this, a number of works have employed multi-view images as the input of networks to learn additional contextual information, namely multi-view learning.
Examples include \citet{conf/MICCAI/chen2018}, \citet{journal/FGCS/yang2020}, and \citet{conf/AIPR/xiao2020} where the features learned from axial, sagittal, and coronal views were combined for the LA cavity segmentation.
Specifically, \citet{conf/MICCAI/chen2018} and \citet{journal/FGCS/yang2020} regarded axial view as the main view due to its finer spatial resolution and extracted information by sequential learning;
and then employed dilated residual learning to extract complementary information from sagittal and coronal views (with lower spatial resolution).
Instead of employing 2D networks, \citet{conf/AIPR/xiao2020} constructed three 3D deep convolutional streams to extract features from the patches of three views, and then fused the features for the LA cavity segmentation.}  

\paragraph{Multi-scale networks}
\reviewerone{}{There exists inconsistency in the sizes of LA anatomical structures such as the PVs among different patients in LGE MRI.
Multi-scale networks are therefore commonly used to learn both local and global features from LGE MRI.
For instance, \citet{journal/JCMARS/du2020} adopted a dual-path structure network with a multi-scale strategy for the LA cavity segmentation from LGE MRI.
\citet{journal/TMI/xiong2018} proposed an AtriaNet consisting of a multi-scale and dual pathway architecture, to capture both local LA tissue geometries and global positional information.
They evaluated their algorithm on 154 LGE MRIs and obtained average Dice scores of $0.940 \pm 0.014$ and $0.942 \pm 0.014$ for the LA epicardium and endocardium, respectively.}

\paragraph{Uncertainty-aware models}
\reviewerone{}{LA structures such as the mitral valve are difficult to segment due to the lack of a clear anatomical border between the LA and the LV.
The ambiguity of the boundary gives rise to uncertainty for the LA cavity segmentation.
\citet{conf/STACOM/yang2018} designed a composite loss to combat uncertainty, and the main idea was to enlarge the gap between background and foreground predictions.
\citet{conf/MICCAI/yu2019} proposed an uncertainty-aware self-ensembling model for semi-supervised LA cavity segmentation.
This is achieved by encouraging the segmentation to be consistent for the same input under different perturbations of the unlabeled data. 
Therefore, they could use abundant unlabeled data for training and obtained similar performance compared to the fully supervised methods using abundant labeled data.}

\subsubsection{\reviewerone{}{Summary of LA cavity segmentation methods}}
\reviewerone{}{In summary, conventional methods generally rely on the information from shape priors or additional paired MRI/ MRA for accurate LA cavity segmentation from LGE MRI.
However, acquiring the auxiliary images requires extra work, and may introduce further errors, i.e., misalignment between LGE MRI and the auxiliary images. 
Recently, with the development of DL and the release of public data, many methods could directly segment the LA cavity from LGE MRI, and achieved promising results.
However, there still exist large errors in the PV and MV regions.
This is mainly due to the small size, the large variability of PVs, including the number, position and orientation of the PVs, and the unclear boundary of MV.
Note that PVs are crucial structures for AF analysis, as scars and ablation gaps are mainly located around PVs after PVI procedures.
To improve the performance of DL-based methods, multi-task learning is effective, and a two-stage network is also a recommended training strategy. 
It is also important to include shape prior or spatial information into the DL-based framework for robust LA cavity segmentation, especially when the size of training dataset is small.
Besides, the accuracy of segmentation was found to be correlated to the image quality of LGE MRI (Pearson's correlation = 0.38\reviewerone{}{, $p$-value = 0.005}) \citep{journal/MedIA/xiong2020}.
It is interesting that the reviewed methods show that 2D and 3D convolutional neural networks (CNNs) had comparable performance, though the target LGE MRI belongs to a 3D image.}

\subsection{LA wall segmentation} \label{method: LA wall seg}

\begin{table*} [t] \center
    \caption{     
     Overview of previously published works on the \textit{LA wall segmentation} from (LGE) MRI and CT.
     MSE: mean square error; A: anterior; P: posterior; Tk: thickness;
     GAC: geodesic active contour; PDE: partial differential equations.
     Note that the evaluation measures and results in \citet{conf/STACOM/inoue2016}, \citet{conf/STACOM/tao2016} and \citet{conf/STACOM/jia2016} are from the benchmark paper \citep{journal/MedIA/karim2018}.
     }
\label{tb:method:LA wall}
{\small
\begin{tabular}{p{2.5cm}|p{2cm}p{3.3cm}p{2.7cm}p{5.7cm}}
\hline
Study    & Data  & Algorithm & Evaluation & Result\\
\hline
\citet{journal/PloS/hsing2014}       &55 LGE MRI    & Manual                          & Tk                            & Tk = $7.0 \pm 1.8$ mm (before ablation)\newline Tk = $10.7 \pm 4.1$ mm (after ablation) \\\hdashline
\citet{journal/MedIA/veni2017}       &72 LGE MRI + 170 Synthetic     & ShapeCut       & ASD, HD, clinical evaluation  & Synthetic: ASD = $0.25 \pm 0.04$ mm; HD = $1.95 \pm 0.38$ mm \newline
                                                                                                                        LGE MRI: ASD = $0.66 \pm 0.14$ mm \newline
                                                                                                                        LGE MRI scar segmentation: MSE = 3.07;  R-square = 0.83\\\hdashline
\citet{journal/JAHA/zhao2017}        &LGE MRI       & Laplace equation                & Tk                            & Tk = $3.7 \pm 1.7$ mm\\\hdashline             \citet{journal/CBE/wang2019}         & 154 LGE MRI + ex vivo data & Convex hull method + coupled PDE       & Tk       & LGE MRI: Tk = 0.4–11.7 mm and median = 3.88 mm\\
\hline\hline
\multirow{3}*{\citet{journal/MedIA/karim2018}}&\multirow{3}*{10 MRI} & Level-set      &\multirow{3}*{Tk, Dice, tissue mass}& Tk = $2.16 \pm 0.58$ mm, Dice = 0.72\\ 
~                                           & ~                      & Region growing & ~                             & Tk = $6.04 \pm 3.63$ mm, Dice = 0.39\\ 
~                                           & ~                      & Watershed      & ~                             & Tk = $3.45 \pm 3.57$ mm, Dice = 0.67\\
\hline\hline
\citet{conf/MI/inoue2014}            &5 enhanced CT & Multi-region segmentation software + manual correction & Tk, visualization     & Tk = 0.5-3.5 mm\\\hdashline
\citet{journal/Europace/bishop2016}  &10 CT         & Morphological operations + Laplace equation             & Tk                    & Errors $\leq 0.2$ mm for Tk of 0.5–5.0 mm\\\hdashline
\citet{conf/STACOM/inoue2016}        &10 CT         & Mesh vertex normal traversal                           & Tk, Dice, tissue mass & Tk = $1.13 \pm 1.02$ mm (A), $1.26 \pm 0.83$ mm (P) \newline Dice = 0.33 (A), 0.39 (P) \\\hdashline   
\citet{conf/STACOM/tao2016}          &10 CT         & Nonlinear intensity transformation + level-set         & Tk, Dice, tissue mass & Tk = $1.34 \pm 0.89$ mm (A), $0.78 \pm 0.41$ mm (P)  \newline Dice = 0.43 (A), 0.21 (P) \\\hdashline                                       
\citet{conf/STACOM/jia2016}          &10 CT         & Region growing + Marker-controlled GAC                 & Tk, Dice, tissue mass & Tk = $0.75 \pm 0.38$ mm (A), $1.46 \pm 1.57$ mm (P)  \newline Dice = 0.30 (A), 0.50 (P)\\
\hline
\end{tabular} }
\end{table*}

\reviewerone{}{To the best of our knowledge, there are limited works reported for automatic LA wall segmentation in the literature, especially from LGE MRI.
Many groups estimated the LA wall from LGE MRI just as an initialization step for the LA scar segmentation \citep{journal/jcmr/Karim2013,journal/MP/yang2018,conf/MICCAI/wu2018}.
These works are not included in this section, as most of them simply dilated the generated LA endocardium by assuming a fixed wall thickness for approximated LA wall segmentation \citep{journal/jcmr/Karim2013}.
However, LA wall thickness varies with positions of the same patient and patients with different gender, age, and disease status \citep{journal/Chest/pan2008}.
With an accurate segmentation result, the wall thickness, which is useful in clinic studies, could be calculated.
For the review of existing techniques of wall thickness measurement, one can refer to Table 1 of the benchmark paper \citep{journal/MedIA/karim2018}.
\textit{Considering the limited number of works reported on LA wall segmentation, in this section we  further review the segmentation on other modalities, including non-enhanced MRI and CT.}
\Leireftb{tb:method:LA wall} summarizes the representative works and results from (LGE) MRI and CT.}

\subsubsection{\reviewerone{}{Conventional methods for LA wall segmentation}}
\paragraph{Morphological operations}
\reviewerone{}{The most straightforward method is to perform morphological operations on the LA endocardium by assuming a fixed wall thickness.
For example, \citet{journal/Europace/bishop2016} adopted morphological operations on the segmented blood pool for wall segmentation from CT.
This method ignores the thickness variation among different LA positions.}

\paragraph{Deformable models}
\reviewerone{}{In contrast, deformable models can dynamically adapt to the changes of wall thickness, and hence obtain more plausible LA wall segmentation results.
For example, \citet{conf/STACOM/tao2016} used the level-set approach to extract the inner and outer LA surface for the final wall segmentation.
\citet{conf/STACOM/jia2016} adopted the region growing method for endocardial segmentation and then utilized Marker-controlled geodesic active contour for the epicardial segmentation.
\citet{journal/MedIA/karim2018} presented the LA wall segmentation and thickness measurement results using three conventional methods, i.e., level-set, region growing, and watershed.
The results showed that level-set performed evidently better than the other two methods; 
region growing generally over-estimated thickness and performed poorly in the wall segmentation task.
They also found that algorithms performed worse in MRI than in CT, which may be due to the fact that the image quality of MRI was generally worse than CT.
However, CT has limited soft tissue contrast, so \citet{conf/STACOM/tao2016} employed nonlinear intensity transformation to enhance the LA wall region in CT.}

\paragraph{Laplace-based solutions}
\reviewerone{}{Laplace-based solutions generate a series of smooth non-intersecting field lines between two boundaries in space and are ideal for simulating the highly variable LA epicardial and endocardial surfaces.
\citet{journal/CBE/wang2019} employed the multi-planar convex hull approach to extract the epicardial and endocardial surfaces, and then used the coupled partial differential equations (PDE) for the wall thickness measurement.
They evaluated their method on both LGE MRI and ex vivo data, and observed that wall thickness values in LGE MRI were more difficult to measure and validate.
Besides, there was a discrepancy in wall thickness measured by ex vivo data and LGE MRI.  
Specifically, the wall thickness values measured from ex vivo data were consistently higher than those measured in LGE MRI.
\citet{journal/JAHA/zhao2017} calculated the wall thickness by solving the Laplace equations on both epicardial and endocardial surfaces.
Despite its prominence, the Laplace-based method still requires explicitly calculating gradient as well as distance trajectories, which are time-consuming and error-prone \citep{journal/CBM/wang2019}. }

\begin{figure}[t]\center
    \includegraphics[width=0.48\textwidth]{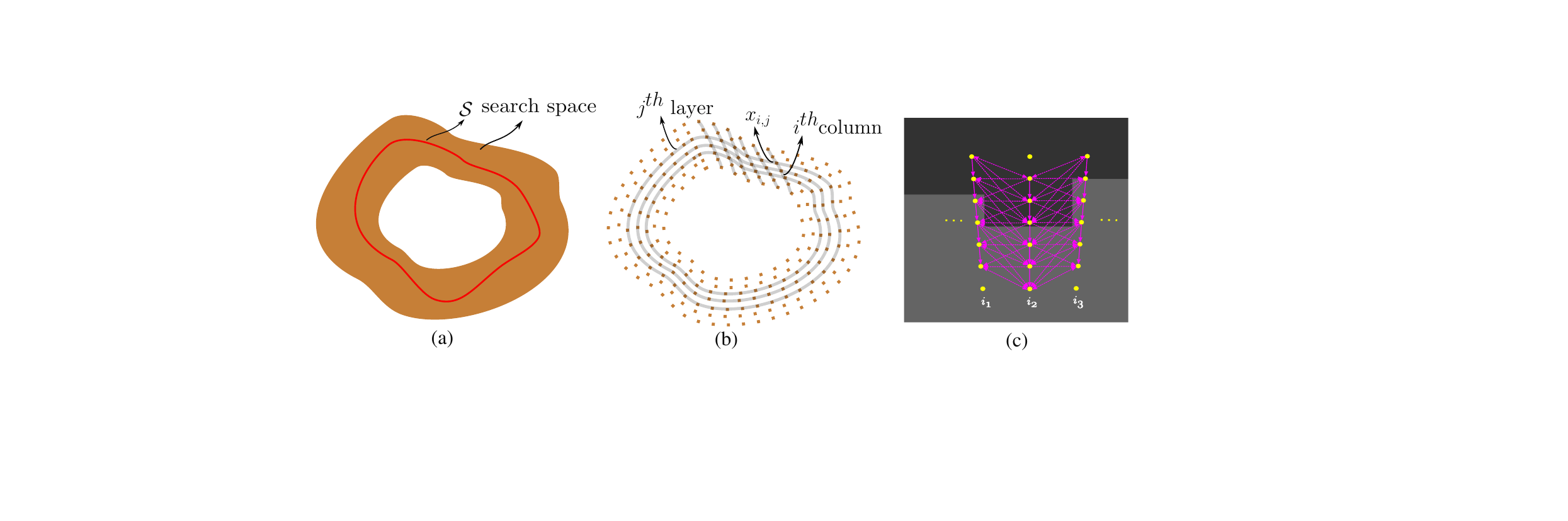}
   \caption{\reviewerone{}{A framework example for the LA wall segmentation from \citet{journal/MedIA/veni2017}. 
   The LA wall segmentation was achieved by predicting the LA epicardium and endocardium respectively, so the task was converted into a surface estimation problem.
   ShapeCut constructed a geometric graph to discretize continuous parameterization of the set of possible surfaces for global optimization.
   (a) continuous parameterization of the surface estimate $S$;
   (b) discrete approximation of the continuous parameterization where the surface estimation performed. Here, each layer maintained a topology similar to the desired surface, and each column ensured that the estimated surface traverses it;
   (c) shape complying properly ordered graph construction.
   Picture modified from \citet{journal/MedIA/veni2017} with permission.} }
\label{fig:method example:LA wall}
\end{figure}

\paragraph{Graph-based methods}
\reviewerone{}{Graph-based methods are promising alternatives.
\citet{journal/MedIA/veni2017} proposed a shape-based generative model namely ShapeCut, to extract epicardial and endocardial surfaces for the LA wall segmentation from LGE MRI, as presented in \Leireffig{fig:method example:LA wall}.
The model could incorporate both local and global shape priors within a maximum-a-posterior estimation framework, and the shape parameters could be optimized via graph-cuts algorithm.
The optimization could be executed in two phases in an iterative manner, i.e., one for multi-surface updates based on multi-column graphs and the other for global shape refinement based on closed forms.
\reviewerone{}{For evaluation, besides directly assessing the LA wall segmentation performance, they also adopted the LA scar segmentation based on their LA wall segmentation for further evaluation.
Specifically, they extracted the scars using thresholding based on both manual and automatic wall segmentations. 
Then, they plotted the fibrosis percentage from manual annotations versus that from automatic ones for each scan. 
They obtained a linear relation with a small error, demonstrating a high overlap between the manual and automatic scarring regions. 
Here, the linear relation error was indicated using the MSE and R-square values.}}

\subsubsection{\reviewerone{}{Summary of LA wall segmentation methods}}
\reviewerone{}{In summary, currently reported works were all based on conventional methods, and no DL-based method has been reported, to the best of our knowledge.  
This could be due to the limited number of relevant public datasets and the large inter- and intra-observer variations of the manual segmentation.
As \citet{journal/MedIA/karim2018} reported, a common error of LA wall segmentation arises from the surrounding tissue such as the neighboring aortic wall.
Improving the image quality may mitigate this problem, and the active contour-based methods with shape constraints and coupled level-set approaches could be helpful.
One of the main applications of LA wall segmentation is to measure wall thickness.
Most of the reported algorithms relied on ruler-based assessments via digital calipers instead of performing a prior segmentation of the LA wall \citep{journal/MedIA/karim2018}.
Several works employed the Laplace equation or PDE to measure wall thickness after achieving the LA wall segmentation.
\citet{journal/MedIA/karim2018} demonstrated that their proposed wall thickness atlas could be effective for thickness prediction in new cases via atlas propagation. 
They constructed a flat thickness map via a surface flattening and unfolding strategy, to compare the mean thickness in each sub-region of the LA wall.
Finally, though CT is a good modality for imaging the thin wall owing to its high resolution, MRI could be effective to assess the wall tissue viability.
Therefore, more attention is expected to the LA wall segmentation from MRI, especially LGE MRI.}


\subsection{LA scar segmentation and quantification} \label{method: LA scar seg/qua}

\begin{table*} [t] \center
    \caption{     
     Summary of previously published works on the (semi-)automatic \textit{LA fibrosis/ scar segmentation and quantification} from LGE MRI.
     $\sim$M Post: $\sim$ months post-ablation; Pre: pre-ablation; 
     XOR: XOR overlap; Percentage: scar percentage; RMSE: root MSE; Volume: total scar volume; ROC: receiver operating characteristic; BER: balanced error rate;
     FCC: fuzzy c-means clustering; MIP: maximum intensity projection; GMM: Gaussian mixture model; SSAE: stacked sparse auto-encoders; MS-CNN: multi-scale CNN; 
     EAM-c: correlation with electroanatomical mapping; FWHM: full-width-at-half-maximum;
     ICC: intraclass correlation coefficients; PCC: Pearson correlation coefficient.
     Here, the asterisk ($^*$) indicates the method employed manual LA wall segmentation.
     }
\label{tb:method:LA scar}
{\small
\begin{tabular}{p{2.5cm}|p{0.4cm}p{1.5cm}p{4.2cm}p{1.3cm}p{3.5cm}p{2cm}}
\hline
Study   & Num  & Pre/ Post & Algorithm & auto? & Evaluation & Dice$_\text{scar}$ \\
\hline
\citet{journal/Circulation/oakes2009}  &81 & Pre              & 2–4 SD                       & semi-auto   &  Percentage, EAM-c             &  N/A\\
\citet{journal/CAE/badger2010}         &144& 3-20M Post       & 3 SD                         & semi-auto   &  Percentage, EAM-c             &  N/A\\
\citet{journal/TBME/knowles2010}       &7  & Post + Pre       & MIP                          & semi-auto   &  Percentage, EAM-c             &  N/A\\
\citet{conf/FIMH/karim2011}            &9  & 6M Post          & Probabilistic intensity model& auto        &  Percentage                    &  N/A\\
\citet{conf/MI/perry2012}              &34 & 3M Post          & K-means clustering           & semi-auto   &  Dice, XOR, percentage         &  $0.807 \pm 0.106$\\\hdashline
\multirow{12}*{\citet{journal/jcmr/Karim2013}}&60 & Post + Pre& Hysteresis threshold         & \multirow{8}*{semi-auto}& \multirow{8}*{Dice, RMSE, volume}&  $0.76^{post}$; $0.37^{pre}$\\
~                                      &60 & Post + Pre       & Region growing + EM-fitting  & ~           &  ~                             &  $0.85^{post}$; $0.22^{pre}$\\ 
~                                      &40 & Post + Pre       & Graph-cuts + FCC             & ~           &  ~                             &  $0.73^{post}$; $0.17^{pre}$\\ 
~                                      &15 & Post             & Active contour + EM-fitting  & ~           &  ~                             &  $0.76^{post}$ \\ 
~                                      &30 & Post + Pre       & Simple threshold$^*$         & ~           &  ~                             &  $0.84^{post}$; $0.48^{pre}$\\
~                                      &60 & Post + Pre       & Graph-cuts                  & ~           &  ~                             &  $0.78^{post}$; $0.30^{pre}$\\ 
~                                      &60 & Post + Pre       & Histogram analysis + threshold$^*$& ~      &  ~                             &  $0.78^{post}$; $0.42^{pre}$ \\ 
~                                      &60 & Post + Pre       & K-means clustering$^*$       & ~           &  ~                             &  $0.72^{post}$; $0.45^{pre}$ \\ \cdashline{2-7}
~                                      &60 & Post + Pre       & 2 SD                         & \multirow{5}*{semi-auto}& \multirow{5}*{Dice, RMSE, volume}&  $0.58^{post}$; $0.24^{pre}$ \\
~                                      &60 & Post + Pre       & 3 SD                         & ~           &  ~                             &  $0.17^{post}$; $0.16^{pre}$ \\
~                                      &60 & Post + Pre       & 4 SD                         & ~           &  ~                             &  $0.14^{post}$; $0.31^{pre}$ \\
~                                      &30 & 1-6M Post        & 6 SD                         & ~           &  ~                             &  $0.35^{post}$ \\
~                                      &60 & Post + Pre       & FWHM                         & ~           &  ~                             &  $0.59^{post}$; $0.05^{pre}$ \\
\hdashline
\citet{journal/TMI/ravanelli2014}      &10 & Pre              & 4 SD                         & semi-auto   &  Dice, EAM-c                   &  $0.600 \pm 0.210$ \\
\citet{journal/TEHM/karim2014}         &15 & 6M Post          & GMM + graph-cuts             & semi-auto   &  Dice, ROC, volume             &  $>0.8$\\
\citet{journal/JMRI/tao2016}           &46 & Pre              & MIP                          & auto        &  Qualitative visualization     &  N/A\\
\citet{journal/MedIA/veni2017}         &72 & Post + Pre       & K-means clustering           & auto        &  Percentage                    &  N/A\\
\citet{journal/MP/yang2018}            &37 & Post             & Super-pixel + SVM            & auto        &  Dice, Acc, Sen, Spe, ROC, BER &  $0.790 \pm 0.050$\\
\citet{conf/MICCAI/wu2018}             &36 & Post             & Multivariate mixture model   & auto        &  Dice, Acc, Sen, Spe           &  $0.556 \pm 0.187$\\
\citet{conf/MI/razeghi2020}            &207& 12M Post  + Pre  & MIP                          & auto        &  ICC, PCC, RMSE                &  N/A \\
\hline \hline
\citet{conf/ISBI/yang2017}             &20 & Post + Pre       & Super-pixel + SSAE           & auto        &  Dice, Acc, Sen, Spe, ROC        &  $0.776 \pm 0.146$\\
\citet{conf/STACOM/lilei2018}          &100& Post + Pre       & Graph-cuts + CNN             & auto        &  Dice, Acc, Sen, Spe             &  $0.566 \pm 0.140 $\\
\citet{conf/MICCAI/chen2018}           &100& Post + Pre       & Multi-view two-task network   & auto        &  Dice, Acc, Sen, Spe, percentage &  $0.78 \pm 0.08$\\
\citet{journal/FGCS/yang2020}          &190& Post + Pre       & Multi-view two-task network   & auto        &  Dice, Acc, Sen, Spe             &  $0.870$\\
\citet{journal/MedIA/li2020}           &58 & 6M Post          & Graph-cuts + MS-CNN          & auto        &  Dice, Acc, Sen, Spe, GDice      &  $0.702 \pm 0.071$\\
\citet{conf/MICCAI/li2020}             &60 & 3-27M Post       & Multi-task network           & auto        &  Dice, Acc, GDice              &  $0.543 \pm 0.097 $\\
\hline
\end{tabular} }\\
\end{table*}

\reviewerone{}{In the literature, a limited number of works have been reported targeting the fully automatic segmentation or quantification of LA scars, probably due to the particular challenge of this task.
Most of the methods require an accurate initial manual segmentation of the LA cavity or LA wall for the following scar classification on the LA wall.
For example, \textit{Left Atrium Fibrosis and Scar Segmentation Challenge} \citep{link/LAScarSeg2012} provided LA cavity labels for participants to develop scar segmentation algorithms.
Eight research teams contributed their methods to this task, including histogram analysis, thresholding, k-means clustering, region-growing with EM-fitting, active contour, and graph-cuts \citep{journal/jcmr/Karim2013}.
The benchmark study showed that semi-automatic methods initialized with manual LA wall segmentation were much more reliable, and performed better than fully automatic approaches \citep{journal/jcmr/Karim2013}.
Currently, the most commonly used approach for the LA scar segmentation is based on thresholding, which is nevertheless sensitive to intensity changes of LGE MRI  \citep{journal/EP/pontecorboli2017}.
\Leireftb{tb:method:LA scar} summarizes all the works, where conventional methods are listed in the upper part and DL-based algorithms are enumerated in the bottom part.}

\subsubsection{\reviewerone{}{Conventional methods for LA scar segmentation and quantification}}

\paragraph{Thresholding}
\reviewerone{}{Thresholding is the most popular method for LA scar segmentation.
The threshold value is normally defined by assuming a fixed standard deviation (SD) above the average intensity value of the normal wall region or blood pool \citep{journal/Circulation/oakes2009,journal/CAE/badger2010,journal/TMI/ravanelli2014}.
For details, one can refer to the survey from \citet{journal/EP/pontecorboli2017}, where different thresholding-based scar segmentation techniques were reviewed and compared.
These methods are easy to implement and intuitive, but also have several disadvantages. 
Firstly, the selection of threshold values is subjective, and the values can differ significantly across various scans, due to the difference of timing from gadolinium administration \citep{journal/TEHM/karim2014,journal/JCMR/chubb2018}.
Secondly, the performance of scar segmentation highly relies on the accuracy of LA or LA wall segmentation that is also challenging, and therefore thresholding based LA scar segmentation was typically achieved via semi-automatic or manual approaches \citep{journal/Circulation/oakes2009,journal/CAE/badger2010}.
The benchmark paper \citep{journal/jcmr/Karim2013} compared eight methods with the full-width-at-half-maximum (FWHM) and $n$-SD methods, and all thresholding methods employed manual LA cavity segmentation as initialization and three of them further utilized manual LA wall segmentation.
In general, all the evaluated eight methods in the benchmark paper outperformed the FWHM and $n$-SD methods.
}

\paragraph{Maximum intensity projection}
\reviewerone{}{Similar to thresholding, maximum intensity projection (MIP) is also a scar quantification scheme that employs scar intensity characteristics.
However, unlike thresholding, MIP is more robust to the inaccurate LA cavity segmentation due to the projection step.
Examples include \citet{journal/TBME/knowles2010} and \citet{journal/JMRI/tao2016}, where projection was performed at $\pm$3 mm and $\pm$2 mm along each normal vector of the LA surface respectively, to consider the potential errors of LA cavity segmentation.
\citet{conf/MI/razeghi2020} also employed MIP for scar segmentation (3 mm externally and 1 mm internally).
Nevertheless, the projection range of MIP must be selected carefully, as it needs to be large enough to extend into the LA myocardium, but not too far to include the intensity of other regions.}

\paragraph{Clustering algorithms}
\reviewerone{}{Considering the complex intensity distribution of LGE MRI, clustering algorithms could be another solution for LA scar segmentation.
This is because clustering can provide a mechanism to statistically separate voxels into groups that are analogous to various tissue types, such as blood pool, healthy wall tissue, and scars.
\citet{conf/MI/perry2012} employed k-means clustering to segment scars from manually segmented LA wall regions.
\citet{journal/MedIA/veni2017} used the same k-means clustering method as \citet{conf/MI/perry2012}, and the LA wall was automatically segmented by their proposed ShapeCut method.
\citet{journal/MP/yang2018} employed super-pixel via a linear iterative clustering algorithm to over segment scars, and then utilized the support vector machine algorithm to classify the over-segmented super-pixels into scarring and normal wall regions.
They scored the image quality into 0 (non-diagnostic), 1 (poor), 2 (fair), 3 (good), and 4 (very good) on a Likert-type scale, according to the level of signal to noise ratio (SNR), appropriate T1, and the existence of navigator beam and ghost artifacts.
Only subjects with image quality $\geq 2$ were selected into their study for evaluation.
\citet{conf/MICCAI/wu2018} combined LGE MRI with anatomical MRI for the scar quantification based on the multivariate mixture model (MvMM) and maximum likelihood estimator (MLE).
They formulated a joint distribution of images using the MvMM \citep{journal/pami/Zhuang2019}, where the registration of the two MRIs and scar segmentation of LGE MRI were performed simultaneously.
Then, the transformation and model parameters were optimized by an iterated conditional model algorithm within the MLE framework. }

\paragraph{Deformable models}
\reviewerone{}{Two deformable models were employed to segment LA scars from LGE MRI, i.e., region growing and active contour with EM-fitting, as reported in \citet{journal/jcmr/Karim2013}.
Among the eight methods mentioned in \citet{journal/jcmr/Karim2013}, region growing with EM-fitting method obtained the best performance on a post-ablation dataset in terms of Dice, even better than those methods that directly employed manual LA wall segmentation for initialization.
For pre-ablation data, the three methods with manual LA wall initialization achieved evidently better Dice compared to the other five methods only with manual LA initialization.
Similar to \citet{journal/MP/yang2018}, \citet{journal/jcmr/Karim2013} classified the LGE MRIs into three types, i.e., good, average, and poor, according to its SNR and contrast ratio (CR) for scars.
They found that most methods had a marginally lower Dice on scans with worse quality, but without statistical significance.
This could be attributed to the minor quality difference and accurate initialization of manual LA cavity segmentation.}

\paragraph{Graph-based methods}
\reviewerone{}{Graph-based methods naturally consider inter-dependencies by introducing links (or edges) between related objects, thus effectively capturing their long-range relatedness.
It may be an effective solution to capture these small and diffuse scars distributed on the LA wall.
\citet{conf/FIMH/karim2011} proposed a probabilistic tissue intensity model which was formulated as a Markov random field and solved using graph-cuts.
In their following work \citep{journal/TEHM/karim2014}, they presented a scar quantification method by combining the scar intensity model priors and Gaussian mixture model (GMM).
Besides, they added constraints via the graph-cuts approach to ensure smoothness and avoided discontinuities in the final scar segmentation.
The proposed method was evaluated on both numerical phantoms and clinical datasets, and demonstrated a good concordance between the automatic results and manual delineations.
Here, numerical phantoms could offer a wide range of variation in scar contrast, which is usually unavailable in clinical datasets.}

\subsubsection{\reviewerone{}{Deep learning-based methods for LA scar segmentation and quantification}}
\reviewerone{}{\citet{conf/ISBI/yang2017} was the first work applying a DL-based classifier for the LA scar segmentation.
Specifically, they used super-pixel over-segmentation for feature extraction, and then adopted a supervised classification step via stacked sparse auto-encoders. 
However, they only used handcrafted intensity features, which provided limited information.
Similar to the DL-based LA cavity segmentation methods, multi-scale, multi-view, and multi-task networks were also employed for LA scar segmentation and quantification.}

\begin{figure}[t]\center
    \includegraphics[width=0.48\textwidth]{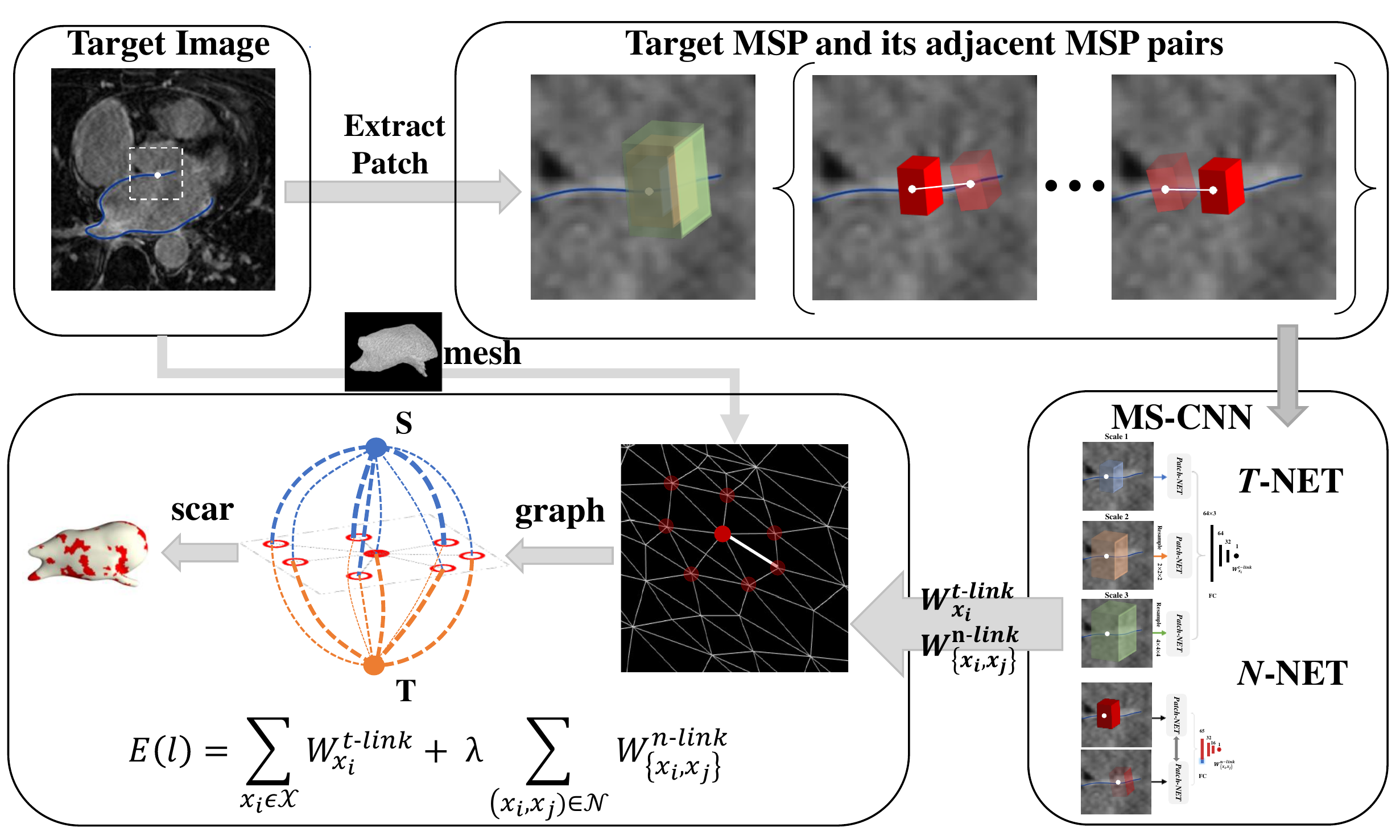}
   \caption{\reviewerone{}{A framework example for the LA scar quantification from \citet{journal/MedIA/li2020}. 
   The scar quantification was performed on the LA surface mesh where a graph was constructed, and the weights of a graph were explicitly learned via an MS-CNN.
   Image adapted from \citet{journal/MedIA/li2020} with permission.} }
\label{fig:method example:LA scar}
\end{figure}

\paragraph{Multi-scale networks}
\reviewerone{}{As \Leireffig{fig:intro:challenges} (d) shows, the surrounding enhanced regions can seriously disrupt the segmentation of scars.
Multi-scale learning could be an effective strategy to alleviate the interference, as it provides both local and global views when learning features of scars.
\citet{conf/STACOM/lilei2018} proposed a hybrid approach utilizing a graph-cuts framework combined with CNNs to predict edge weights of the graph for the automatic scar segmentation.
They extended their work by introducing multi-scale CNN (MS-CNN) to learn local and global features simultaneously \citep{journal/MedIA/li2020}, as presented in \Leireffig{fig:method example:LA scar}.
The experimental results showed that the multi-scale learning scheme (number of scales = 3) improved the performance when compared with a single scale (Dice$_\text{scar}$: $0.702 \pm 0.071$ vs. $0.677 \pm 0.070$).
Besides, the scheme is also less dependent on an accurate LA cavity segmentation, which makes it more robust.
A major limitation of this study was the lack of an end-to-end training style, as the framework was split into three sub-tasks, i.e., LA cavity segmentation as an initialization, feature learning via the MS-CNN, and optimization based on graph-cuts.
This indicated the limitation of multi-scale patch strategies, which resulted in an expensive time and space complexity and an infeasible end-to-end training on the whole graph.}

\paragraph{Multi-task/ multi-view networks}
\reviewerone{}{To achieve end-to-end optimization, multi-task learning is desired.
\citet{conf/MICCAI/li2020} developed a new framework where LA cavity segmentation, scar projection onto the LA surface, and scar quantification are performed simultaneously in an end-to-end fashion based on a multi-task network.
In this framework, they proposed a shape attention (SA) mechanism by an implicit surface projection, to utilize the inherent spatial relationship between the LA cavity and scars.
The mechanism also alleviated the class-imbalance problem in the scar quantification and proved to be effective in the ablation study.
Similarly, \citet{conf/MICCAI/chen2018} and \citet{journal/FGCS/yang2020} adopted multi-task learning for simultaneous LA and scar segmentation, but the spatial relationship between the two regions was not explicitly learned in their works.
Moreover, as mentioned in Section \ref{LA method: multi-view network}, they employed multiple views as the input of multi-task networks.}

\subsubsection{\reviewerone{}{Summary of LA scar segmentation and quantification methods}}

\reviewerone{}{In summary, scar segmentation/ quantification from LGE MRI remains an open problem.
Most methods relied on interactive correction/ manual initialization, or on accurate initial estimation of LA wall segmentation for following application of thresholding.
These semi-automatic approaches generally obtained high accuracies in terms of Dice scores.
Compared to the conventional automatic methods, DL-based algorithms could obtain better performance.
However, DL-based models could have limited model generalization ability. 
In general, pre-ablation data with fibrosis is more challenging to segment than post-ablation data with scars.
This may be attributed to the fact that fibrosis appears more diffusely compared to post-ablation scars \citep{journal/jcmr/Karim2013}.
In addition, it is difficult to differentiate the native fibrosis and post-ablation scars for longstanding persistent AF patients \citep{conf/MI/yang2017}.
One major challenge for scar segmentation/ quantification is the artifacts from the boundary regions, such as the right atrial (RA) wall and aorta wall.
A good initialization, i.e., accurate LA or LA wall segmentation, could be helpful to counteract this problem.
\citet{journal/MedIA/li2020} tried to reduce the dependence on accurate LA cavity segmentation via projection and MS-CNN, while \citet{conf/MICCAI/li2020} introduced a distance-based spatial encoding loss for training a deep neural network to learn the spatial information of scars around the LA boundary}.
Another challenge arises from the imaging, including poor image quality and data-mismatch issues in DL-based methods.
Therefore, a more consistent and standard image acquisition protocol is highly required. 
Alternatively, domain generalization algorithms need to be considered to improve the model generalization ability across different sites or on unseen datasets \citep{conf/STACOM/li2020,journal/TMI/campello2021}.

\begin{table*} [t] \center
    \caption{     
     Summary of previously published works on the (semi-)automatic \textit{ablation gap quantification} from LGE MRI.
     \# gaps: mean number of gaps; GL: gap length; IIR: image intensity ratio; NAUC: normalized area under the curve;
     RSPV: right superior PV; LIPV: left inferior PV; LSPV: left superior PV.
     }
\label{tb:method:LA gap}
{\small
\begin{tabular}{p{2.8cm}|p{0.4cm}p{3cm}p{2.5cm}p{7.4cm}}
\hline
Study   & Num  & Algorithm & Evaluation & Results \& Main findings  \\  
\hline
\citet{journal/CAE/badger2010}  &144&3 SD for scar segmentation + visually detect gap   &Visual, EAM-c                & Significant relationship between gaps and recurrence;\newline
                                                                                                                        Achieving complete circumferential lesions around the PV is difficult.\\\hline
\citet{journal/CAR/ranjan2012}  &12 &Measurement tool in \href{https://www.osirix-viewer.com/}{\textit{Osirix}}                         &GL, pathology correlation    & The correlation coefficient ($R^2$) between the GL identified by LGE MRI and the gross pathology was 0.95; \newline                                                                                                                GL = 1.0 mm (via gross pathology) and GL = 1.4 mm (via LGE MRI); \newline
                                                                                                                        Real time MRI system can be used to identify gaps.\\\hline
\citet{journal/JACC/bisbal2014} &50 &Manual LA wall + MIP                               &GL, \# gaps, EAM-c           & \# gaps = 4.4/patient; \# gaps = $1.27 \pm 0.41 $/PV \newline
                                                                                                                        Median GL = $13.33 \pm 5.8 $ mm/gap;\newline
                                                                                                                        Position of highest \# gaps: RSPV (=1.53); \newline
                                                                                                                        Position of fewest \# gaps: LIPV (=0.67);\newline
                                                                                                                        The majority of patients (73.3\%) had gaps in all PVs.\newline
                                                                                                                        LGE MRI may identify non-conducting gaps that could be related to later recurrences.\\\hline
\citet{journal/CAE/harrison2015}&20 &Custom-written software                            &Visual, EAM-c                & Weak point-by-point relationship ($R^2$=0.57) between scars and endocardial voltage in patients undergoing repeat LA ablation;\newline                                                                                                    The mean voltage within scar region is lower than that within normal wall region.\\\hline
\citet{journal/CAE/linhart2018} &94 &IIR for scar segmentation + gap is defined as discontinuation of the ablation line by $\leq3$ mm   
                                                                                        &GL, \# gaps, EAM-c           & \# gaps = 5.4/patient; Mean GL = 7.3 mm/gap; \newline
                                                                                                                        90 out of 94 patients (96\%) had at least 1 anatomic gap;\newline
                                                                                                                        Anatomic gaps are frequently detected in LGE MRI at 3 months after first PVI;\newline
                                                                                                                        An increase of 10\% relative GL increased the likelihood of AF recurrence by 16\%.;\newline
                                                                                                                        The total relative GL was significantly associated with recurrence instead of \# gaps.\\\hline
\citet{journal/JA/mishima2019}  &10 &2 SD for scar segmentation + visually detect gap   &GL, \# gaps, EAM-c           & Mean GL = $11.6 \pm 3.9 $ mm/gap;\newline
                                                                                                                        \# gaps = 1.6/patient (1st ablation); \newline
                                                                                                                        \# gaps = 1.4/patient (2nd ablation); \newline
                                                                                                                        Position of highest \# gaps: RSPV (=2); \newline
                                                                                                                        Position of fewest \# gaps: LIPV (=0);\newline
                                                                                                                        The location of electrical gaps are well matched to that on the LGE MRI.\\ \hline
\citet{journal/MedIA/nunez2019} &50 &Graph-based method                                 &GL, \# gaps, RGM    & Position of highest \# gaps: LSPV (=1.73); \newline
                                                                                                                        Position of fewest \# gaps: LIPV (=1.16);\newline
                                                                                                                        No significant differences between left and right PVs;\newline
                                                                                                                        No significant relationship between gaps and recurrence.\\
\hline
\end{tabular} }\\
\end{table*}

\subsection{LA ablation gap quantification} \label{method: LA gap qua}

\reviewerone{}{Gaps around PVs can be classified into electrical/ conduction gaps and anatomical ablation gaps. 
Conduction gaps refer to the electrical reconnection regions with high voltages in the electroanatomical mapping (EAM), and they can be detected using intra-cavitary catheters during a redo procedure.
Ablation gaps indicate the gaps of healthy tissue in the (ideally continuous) scars, which are typically identified by LGE MRI.
Therefore, in this section, we only focus on the developed methods to quantify ablation gaps from LGE MRI.
\reviewerone{}{Note that the ablation gaps do not belong to the inherent structure of the LA, but instead are ``gaps'' left during the LA ablation procedure.}
\Leireftb{tb:method:LA gap} summarizes representative (semi-)automatic LA ablation gap quantification methods, results, and main findings.}

\subsubsection{\reviewerone{}{Conventional methods for LA gap quantification}}

\paragraph{Visual detection}
\reviewerone{}{To the best of our knowledge, most of the methods reported in the literature relied on visual inspection, which could result in biased estimations of gap characteristics, such as the number, length, and position of gaps.
For instance, \citet{journal/CAE/badger2010} and \citet{journal/JA/mishima2019} both employed thresholding for the scar segmentation and then detected ablation gaps visually.
Moreover, as ablation gaps are highly correlated with scars, there is a certain overlap for quantification methods of scars and ablation gaps, such as MIP and thresholding.
\citet{journal/JACC/bisbal2014} manually segmented the LA wall for an accurate initialization and then adopted MIP for the scar and gap classification.
\citet{journal/CAE/linhart2018} used the image intensity ratio as a threshold for LA scar segmentation and defined the gaps as the discontinued ablation line $\leq3$ mm.
Several software packages were also employed for ablation gap quantification, such as \href{https://www.osirix-viewer.com/}{\textit{Osirix}} \citep{journal/CAR/ranjan2012} and Custom-written software  \citep{journal/CAE/harrison2015}.}

\begin{figure}[t]\center
    \includegraphics[width=0.48\textwidth]{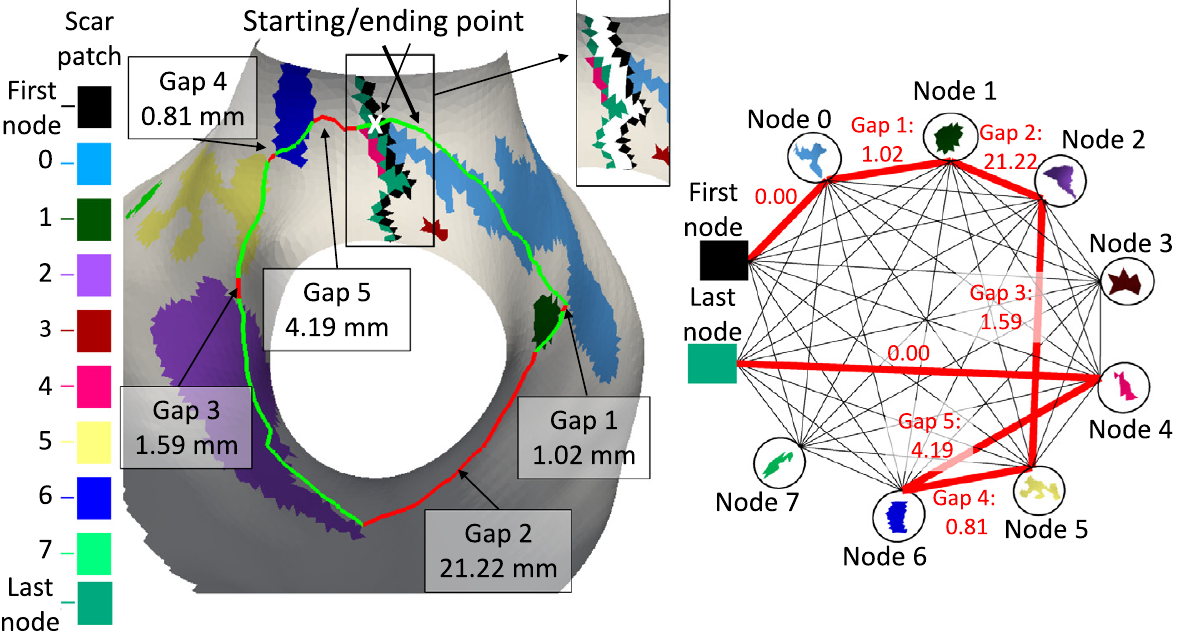}
   \caption{\reviewerone{}{A framework example for the LA gap quantification from \citet{journal/MedIA/nunez2019}. 
   The gap quantification was performed through minimum path search in a graph where every node referred to a scarring patch and the edges denoted the geodesic distances between patches.
   Image adapted from \citet{journal/MedIA/nunez2019} with permission.} }
\label{fig:method example:LA gap}
\end{figure}

\paragraph{Graph-based methods}
\reviewerone{}{Recently, \citet{journal/MedIA/nunez2019} proposed a reproducible framework for semi-automatic gap quantification using a graph-based method, as presented in \Leireffig{fig:method example:LA gap}.
One can see that the gap quantification was performed via minimum path search in a graph where each node was a scarring patch, and the edges denoted the geodesic distances between patches.
They proposed a quantitative measure to estimate the percentage of gaps around a vein, namely the relative gap measure.
One major limitation of this work was that a fixed regional parcellation was assumed, i.e., four-PV configuration in the LA, but actually only around 70\% of LA have four PVs \citep{journal/JCDR/prasanna2014}.}

\subsubsection{\reviewerone{}{Summary of LA ablation gap quantification methods}}
\reviewerone{}{It is considered difficult to achieve complete circumferential lesions, so the majority of patients have gaps after ablation \citep{journal/CAE/badger2010,journal/JACC/bisbal2014,journal/CAE/linhart2018}.
The most common locations appearing gaps are the area between the left superior PV (LSPV) and the LAA.
This may be due to the presence of a thicker myocardium in this area, which leads to non-transmural lesions \citep{journal/ACD/galand2016}.
In \citet{journal/JACC/bisbal2014} and \citet{journal/JA/mishima2019}, the largest number of gaps occurred in right superior PV (RSPV) was reported;
while in \citet{journal/MedIA/nunez2019} it appeared in LSPV.
In contrast, the fewest of gaps occurred consistently in the left inferior PV (LIPV)
 \citep{journal/JACC/bisbal2014,journal/JA/mishima2019,journal/MedIA/nunez2019}.
The different distributions of gaps in different PV positions could be attributed to the differences in imaging and limited accuracy of scar segmentation in these regions.}

\reviewerone{}{The relationship between electrical gaps of EAM and anatomical gaps of LGE MRI is still unclear.
\citet{journal/JA/mishima2019} found that the location of electrical gaps was well matched to that of the detected ablation gaps from LGE MRI.
However, \citet{journal/CAE/harrison2015} claimed a weak point-by-point relationship between scars and EAM in the patients with repeated LA ablation.
Besides, the relationship between ablation gaps and AF recurrence is also controversial, with positive answers \citep{journal/JACC/peters2009,journal/HR/taclas2010,journal/CAE/badger2010,journal/JACC/bisbal2014,journal/CAE/linhart2018} but also negative conclusions \citep{journal/HR/spragg2012,journal/CCIP/harrison2015,journal/MedIA/nunez2019}.  
These are partially due to the lack of an objective and consistent method for ablation gap quantification, primarily depending on visual observation.
The task has not been properly addressed in the literature, and research on this is still in an early stage.}

\subsection{\reviewerone{}{Image computing and analysis on the LA LGE MRI}} \label{method:LA analysis}

\reviewerone{}{So far, we have presented and discussed the recent progress in LA LGE MRI computing.
\Leireftb{tb:method:summary} summarizes the various properties of different targets with corresponding potential processing schemes.
The LA cavity is a relatively large target but with variable shapes;
the LA wall is equivalent to two surfaces with extremely small and inconsistent distance;
and the LA scars/ ablation gaps belong to small, discrete, and space-constrained (scars and ablation gaps are localized at the LA wall) targets with distinct features.  
Most of the methods summarized here are customized to the corresponding attributes and challenges of each task.
For example, due to the variable shapes of the LA cavity, many atlas-based methods were proposed to incorporate the shape priors.
Auxiliary images, uncertainty-aware, and coarse-to-fine training schemes are also beneficial for LA cavity segmentation.
Due to the properties of the LA wall, variants of deformable models were employed, such as coupled level-set, region growing, and watershed algorithms. 
With an accurate LA initialization, it is straightforward yet effective to adopt thresholding for the scar segmentation, as scarring regions are enhanced in intensity compared to the healthy wall. 
Moreover, due to the thin wall, some researchers proposed to project the scars onto the LA surface  ignoring the wall thickness for scar quantification.}

\begin{table*} [t] \center
    \caption{\reviewerone{}{Summary of current computing methods on the LA LGE MRI for AF analysis.} }
\label{tb:method:summary}
{\small
\begin{tabular}{p{1.25cm}|p{7cm}p{8cm}}
\hline
Target    & Unique characteristic & Potential processing strategies \& future perspective \\ 
\hline
LA cavity & large shape variability  & introducing shape prior; combining complementary information from other paired modalities; uncertainty-aware schemes; coarse-to-fine training \\
\hdashline
LA wall   & thin thickness; irregular opening; varying thickness across the wall & level-sets; ShapeCut and its possible future version combining DL-based feature extraction  \\
\hdashline
LA scars  & small; diffusion; spatial constrained; intensity-related; corresponding to the low voltage regions of EAM & multi-scale learning schemes; surface projection; multi-task learning schemes; thresholding based on an accurate initialization \\ 
\hdashline
LA gaps   & without an unified definition; uncertain number of PVs & quantitative instead of visual qualitative quantification  \\
\hline
\end{tabular} }\\
\end{table*}

\reviewerone{}{Nevertheless, there is a certain overlap in these reviewed approaches, mainly as the four tasks are coherent and share similar challenges (please refer to Sec 1.2 of the manuscript for the challenges of each task).
Among the \textit{conventional methods}, several classical algorithms were commonly employed, such as graph-based methods, deformable models, and clustering algorithms.
For example, \Leireffig{fig:method example:LA wall}, \Leireffig{fig:method example:LA scar}, and \Leireffig{fig:method example:LA gap} present the graph-based methods for LA wall segmentation, LA scar quantification, and LA gap quantification, respectively.
It is evident that for different tasks the graphs were constructed in different styles.
Specifically, for LA wall segmentation a graph was represented by a set of columns and a neighborhood structure among adjacent columns for a multi-surface update, namely a multi-column graph.
For LA scar quantification, a graph was designed on the LA surface mesh, and the graph weights were learned by MS-CNN.
For LA gap quantification, the scar patches were regarded as the nodes of a graph, and the geodesic distances between patches were denoted as the edges.
Among the \textit{DL-based methods}, there are several commonalities for LA cavity and scar segmentation/ quantification, which can be categorized into three kinds,
(1) alleviating the class imbalance problem via pre-processing, a two-stage pipeline, or weighted sampling; 
(2) improving the robustness of networks via multi-scale learning, multi-task learning, or multi-view feature fusion;
(3) forcing the network to generate more plausible segmentation results by incorporating shape priors, applying anatomical constraints, or introducing uncertainty maps.
It is worthwhile to highlight that for LA cavity and scar segmentation/ quantification, leveraging spatial relationship of LA cavity and scars via simultaneous optimization has been explored and shown to be beneficial for improving the accuracy.}

\reviewertwo{}{There are apparent trade-offs between conventional and DL-based algorithms.
Conventional approaches are transparent and well-established, while DL has potential of higher precision and versatility but with the cost of an enormous amount of  data and computing resources \citep{journal/SIC/o2019}.   
Therefore, it is interesting to explore hybrid approaches combining the advantages of them.
Several works have demonstrated their benefits for LA LGE MRI computing. 
For example, for LA cavity segmentation, \citet{conf/STACOM/borra2018} utilized Otsu’s algorithm to extract ROI and then performed segmentation on the ROI via U-Net.
\citet{journal/MedIA/li2021} employed the conventional distance transform maps to incorporate continuous spatial information of the target label.
The limited receptive view and spatial awareness in the standard CNN-based methods could lead to a noisy segmentation, especially for the target with highly variable shapes, such as LA.  
Their results showed the effectiveness of distance transform maps in the DL-based framework removing the noisy patches of the segmentation. 
Statistical shape models (SSMs) can be a promising alternative to combine CNN with prior knowledge of anatomical shapes for the LA cavity segmentation \citep{journal/MedIA/ambellan2019}.
Recently, DL-based cross-modality MAS frameworks are promising for the left ventricle (LV) myocardial segmentation \citep{conf/MICCAI/ding2020}, and could be extended for the LA cavity segmentation, especially when the additional paired modalities are available.
For LA scar quantification, \citet{journal/MedIA/li2020} combined the conventional graph-cuts algorithm and MS-CNN (LearnGC) for hybrid representations of structural and local features.
They employed MS-CNN to learn multi-scale features of patches corresponding to nodes on the graph and obtained better results than conventional graph-cuts algorithms which were based on hand-crafted features. 
For LA wall segmentation and gap quantification, no DL-based method has been reported, to the best of our knowledge. 
However, the  conventional ShapeCut algorithm proposed by \citet{journal/MedIA/veni2017} can be adapted for such application, by extracting features from the intensity profiles via CNN for more accurate LA wall segmentation.
Similar schemes can be employed on the proposed graph-based method for LA gap quantification \citep{journal/MedIA/nunez2019}. 
Moreover, the utility of level-set for LA wall segmentation has been proven \citep{journal/MedIA/karim2018}, and the combination of DL and level-set for the LV segmentation obtained accurate results with small training sets \citep{journal/MedIA/ngo2017}.
Therefore, such combination and hybrid approaches are expected and should be further explored in the near future.
}

\begin{table*}\center
\caption{Summary of the public datasets whose research targets are AF patients or include AF patients. 
        Here, the star ($^\star$) indicates that the data is acquired from multiple centers and vendors.
        bSSFP: balanced steady-state free precession.
        }
\label{tb:intro:public dataset}
{\small
\begin{tabular}{p{3cm}|p{0.5cm}p{3.2cm}p{2.5cm}p{7cm}}
\hline
Source	   &  Year & Data & Target & Name \& Hyperlink \\
\hline
\citet{link/CARMA2012}          & 2012 & 155 LGE MRI + 3D MRI	      & LA cavity, LA scar  & \href{http://insight-journal.org/midas/\%20collection/view/197}{\textit{CARMA, University of Utah}} \\ 
\citet{journal/jcmr/Karim2013}	& 2012 & 60 LGE MRI$^\star$		      & LA scar      & \href{http://www.cardiacatlas.org/challenges/left-atrium-fibrosis-and-scar-segmentation-challenge/}{\textit{Left Atrium Fibrosis and Scar Segmentation Challenge}} \\
\citet{journal/tmi/Tobon2015}	& 2013 & 30 CT, 30 bSSFP MRI          & LA cavity	       &\href{https://www.cardiacatlas.org/challenges/left-atrium-segmentation-challenge/}{\textit{Left Atrium Segmentation Challenge}} \\
\citet{journal/MedIA/karim2018}	& 2016 & 10 CT, 10 black-blood MRI    & LA wall    &\href{https://www.doc.ic.ac.uk/\~rkarim/la\_lv\_framework/wall/datasets.html}{\textit{Left Atrial Wall Thickness Challenge}}\\
\citet{journal/MedIA/zhuang2019}& 2017 & 60 CT, 60 bSSFP MRI$^\star$  & Whole heart including LA cavity &\href{http://www.sdspeople.fudan.edu.cn/zhuangxiahai/0/mmwhs/}{\textit{Multi-Modality Whole Heart Segmentation Challenge}}\\
\citet{journal/MedIA/xiong2020}	& 2018 & 150 LGE MRI	              & LA cavity	       &\href{http://atriaseg2018.cardiacatlas.org/data/}{\textit{Atrial Segmentation Challenge}}\\
\hline
\end{tabular}}\\
\end{table*}


\section{Data and evaluation measures} \label{evaluation issue}

Validation work not only reveals the performance and limitations of a proposed method, but also clarifies the scope of its application \citep{journal/TMI/jannin2006}.
Hence, it is essential to validate an algorithm before applying it to a clinical setting.
This section examines and analyzes the validation methods used for each aforementioned task in the literature, including the data and performance measures.
We also focus on the evaluation of clinically relevant measures, besides the evaluation of computing accuracy of the algorithms. 

\subsection{\reviewerfour{}{Public AF related datasets}} \label{public dataset}
\reviewerfour{}{Several challenge events have been organized in recent years at international conferences such as ISBI (International Symposium on Biomedical Imaging) and MICCAI (Medical Image Computing and Computer-Assisted Interventions), with corresponding public datasets released.
For example, Zhuang et al. organized the \textit{Multi-Modality Whole Heart Segmentation Challenge}, in conjunction with STACOM'17 and MICCAI'17.
They provided 120 multi-modality images covering a wide range of cardiac diseases, such as AF, myocardial infarction, and congenital heart disease \citep{journal/MedIA/zhuang2019}.
Ten algorithms for CT data and eleven methods for MRI data have been evaluated, and most of the submitted algorithms were DL-based.
The evaluated results showed that the LA cavity segmentation of AF patients was particularly more accurate compared to other categories of patients.
Moreover, public datasets were released along with the challenge events focusing on a specific anatomical structure instead of the whole heart.
\Leireftb{tb:intro:public dataset} summarizes the public AF-related events and datasets with corresponding download links.}

\reviewerfour{}{For LA cavity segmentation, Tobon-Gomez et al. organized the \textit{Left Atrium Segmentation Challenge}, in conjunction with STACOM'13 and MICCAI'13.
They offered a dataset including 30 CT and 30 MRIs with the manual LA cavity segmentation and presented the results of nine algorithms for CT and eight for MRI \citep{journal/tmi/Tobon2015}.
Their results showed that the methodologies that combined statistical models with region-growing were the most suitable for the target task.
Zhao et al. organized the \textit{Atrial Segmentation Challenge}, in conjunction with STACOM'18 and MICCAI'18.
They provided 150 LGE MRIs with manual LA cavity segmentation generated from three experts, and the data covered both pre- and post-ablation images \citep{journal/MedIA/xiong2020}.
To explore the quality of the dataset, they calculated three measures, i.e., SNR, CR, and heterogeneity, which were in agreement.
The quality measurements showed that less than 15\% of the data had high quality (SNR$>$3), 70\% had medium quality (SNR = 1$\sim$3), and over 15\% was of low quality (SNR$<$1).
In total, 27 teams contributed to the automatic LA cavity segmentation, and most of the methods were DL-based except for two MAS methods. 
The results showed that two-stage CNNs achieved superior results than other single CNN methods and conventional methods.
This challenge event provided a significant step towards much-improved segmentation methods for the LA cavity segmentation of LGE MRI.}

\reviewerfour{}{For LA wall segmentation, Karim et al. organized the \textit{Left Atrial Wall Thickness Challenge}, in conjunction with STACOM'16 and MICCAI'16.
The released images consisted of 10 CT and 10 MRIs of healthy and diseased subjects with manual LA wall segmentation. 
Only two of the three participants contributed to the automatic segmentation of the CT data, but no work on the MRI data was reported \citep{journal/MedIA/karim2018}.
The limited number of submitted algorithms generally performed poorly compared to the inter-observer variability, which revealed the difficulty of the wall segmentation task.
\citet{link/LAseg2018} and \citet{link/CARMA2012} released a public LGE MRI dataset with LA wall segmentation.
This segmentation was however generated using the morphological (dilation) operation from the LA cavity manual segmentation.
}

\reviewerfour{}{For LA scar segmentation, Karim et al. organized the \textit{Left Atrium Fibrosis and Scar Segmentation Challenge} at ISBI 2012.
They provided 60 multi-center and multi-vendor LGE MRIs with manual labels of both LA and scars, and summarized the submitted algorithms from seven institutions in \citet{journal/jcmr/Karim2013}.
To the best of our knowledge, no public dataset for gap quantification and evaluation has been reported.}


\subsection{Evaluation measures} \label{evaluation measures}
\begin{figure*}[t]\center
    \includegraphics[width=0.64\textwidth]{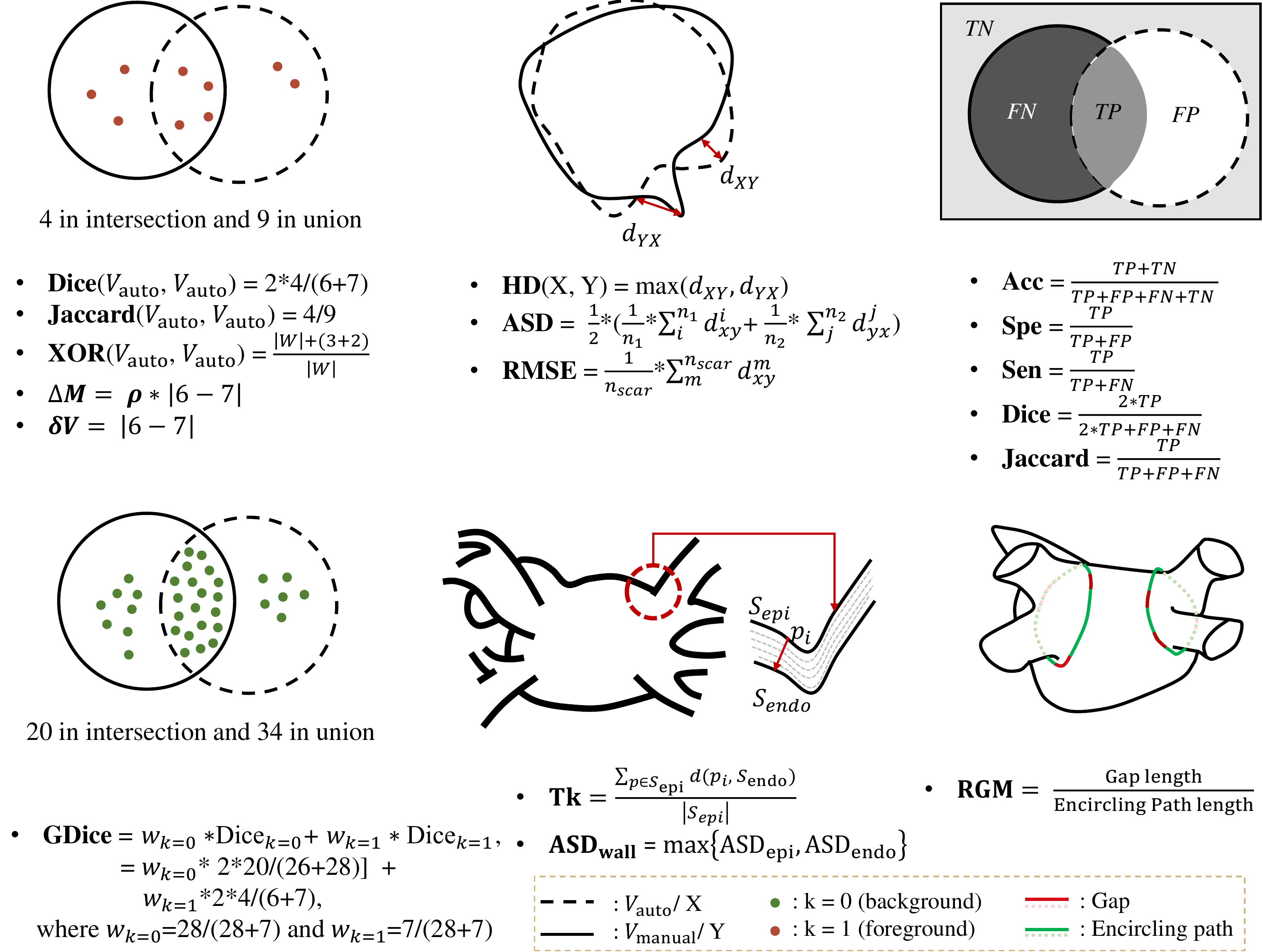}
   \caption{\reviewerfour{}{Sketch map of measures employed in LA LGE MRI computing.} }
\label{fig:evaluation:LA}
\end{figure*}

\reviewerfour{}{The methods are evaluated in different ways for different tasks in the literature.
However, all the measures are generally designed based on the idea of comparing automatic segmentation results with reference segmentations.
In this section, we summarize common measures employed in each LA computing task.
The reader is referred to \Leireffig{fig:evaluation:LA} for an illustration of each evaluation measure listed below.}

\subsubsection{LA cavity measures}
For assessing the performance of LA cavity segmentation, a range of different measures have been explored, as shown in \Leireftb{tb:method:LA}.
The most widely used measures include the Dice coefficient/ score, Jaccard index, HD, and average surface distance (ASD). They are defined as follows,
\begin{equation}
  \mathrm{Dice}(V_{\mathrm{auto}}, V_{\mathrm{manual}}) = \frac{2\left|V_{\mathrm{auto}} \cap V_{\mathrm{manual}}\right|}{\left|V_{\mathrm{auto}}\right|+\left|V_{\mathrm{manual}}\right|},
\end{equation}
\begin{equation}
  \mathrm{Jaccard}(V_{\mathrm{auto}}, V_{\mathrm{manual}}) = \frac{\left|V_{\mathrm{auto}} \cap V_{\mathrm{manual}}\right|}{\left|V_{\mathrm{auto}}  \cup V_{\mathrm{manual}}\right|},
\end{equation}
\begin{equation}
  \mathrm{HD}(X, Y)=\max \Big[\sup _{x \in X} \inf _{y \in Y} d(x, y), \sup _{y \in Y} \inf _{x \in X} d(x, y)\Big],
\end{equation}
and
\begin{equation}
    \mathrm{ASD}(X, Y)=\frac{1}{2}\left(\frac{\sum_{x \in X} \min _{y \in Y}d(x, y)}{\sum_{x \in X} 1}+\frac{\sum_{y \in Y} \min _{x \in X}d(x, y)}{\sum_{y \in Y} 1}\right),
\end{equation}
where $V_{\mathrm{manual}}$ and $V_{\mathrm{auto}}$ denote the set of pixels in the manual and automatic segmentation, respectively;
$X$ and $Y$ represent two sets of contour points;
$d(x, y)$ indicates the Euclidean distance between the two points $x$ and $y$;
and $|\cdot|$ refers to the number of pixels in set $V$.
Dice and Jaccard are selected for volumetric overlap measurement, where Jacquard index can be more sensible and severe upon small variation compared to Dice \citep{conf/STACOM/jamart2019}.
ASD and HD are used to evaluate the shape and contour accuracy of the object of interest.
ASD calculates the average of the distances between all pairs of pixels between two surfaces.
HD calculates the largest error distance of the 3D segmentation defined for a prediction of the target.
Therefore, HD can further measure the existence of outliers, and sometimes 95\% HD will be used to eliminate the influence of a small subset of outliers.

In addition, three statistical measurements are employed, i.e., Accuracy (Acc), Specificity (Spe), and Sensitivity (Sen), defined as follows,
\begin{equation}
    \mathrm{Acc}=\frac{TP+TN}{TP+FP+FN+TN}, 
\end{equation}
\begin{equation}
    \mathrm{Spe}=\frac{TN}{TN+FP},
\end{equation}
and
\begin{equation}
    \mathrm{Sen}=\frac{TP}{TP+FN}, 
\end{equation}
where $TP$, $TN$, $FN$, and $FP$ stand for the number of true positives, true negatives, false negatives, and false positives, respectively. 
Acc represents the proportion of true results (both $TP$ and $TN$) among the total number of cases examined.
Spe and Sen are used to reflect the success of the algorithm for the foreground and the background segmentation, respectively.
Besides, the diameter and volume error calculations are used to assess the medical relevance of the automatic reconstructed LA volumes in the clinic.

\subsubsection{LA wall measures} \label{wall_measure}
For the LA wall segmentation, wall thickness and Dice are currently the most commonly used measures.
The thickness (Tk) of the LA wall can be calculated by averaging the thickness over each pixel $p_{i} \in \mathrm{S_{epi}}$ from the epicardium $\mathrm{S_{epi}}$ to the endocardium $\mathrm{S_{endo}}$, and therefore is defined as,
\begin{equation}
    \mathrm{Tk}=\frac{\sum_{p_{i} \in \mathrm{S_{epi}}} d\left(p_{i}, \mathrm{S_{endo}}\right)}{|\mathrm{S_{epi}}|}.
\end{equation}
Actually, when the object size is much smaller than the background (as in the case of the LA wall), overlap-based metrics based on the four overlap cardinalities (TP, TN, FP, FN) are generally inappropriate \citep{journal/BMC/taha2015}.
This is because they will provide the same metric value, regardless of the distance between two non-overlapping regions evaluated, ultimately affecting the objectivity in precision.
Therefore, both Dice and Jaccard are not suitable since they can also be represented as,
\begin{equation}
  \mathrm{Dice} = \frac{2TP}{2TP+FP+FN},
\end{equation}
\begin{equation}
  \mathrm{Jaccard} = \frac{TP}{TP+FP+FN}. 
\end{equation}
In this case, distance-based metrics are recommended, as they consider the precision and accuracy of both the shape and local alignment of segmented regions.
Apart from its small size, the LA wall is also accompanied by adjacent PV structures, which also exhibit large inter-observer variation and could be regarded as outliers.
Compared to HD which is sensitive to outliers, ASD is a better option for LA wall quantitative assessment.
As the LA wall segmentation involves the two surfaces, i.e., the epicardium and endocardium, the ASD of the LA wall is defined as,
\begin{equation}
  \mathrm{ASD_{wall}} = \max \left\{\mathrm{ASD_{epi}}, \mathrm{ASD_{endo}} \right\}.
\end{equation}
Apart from these measurements, tissue mass and clinical evaluation are also employed for the evaluation of LA wall segmentation.
The tissue mass $M$ is designed to predict the volume error, and the difference in mass is defined as,
\begin{equation}
  \Delta M = \rho \times|V-\hat{V}|,
\end{equation}
where $\rho=1.053$ g/ml \citep{journal/AJPHCP/vinnakota2004} is the average wall tissue density, and $V$ and $\hat{V}$ refer to the reference and predicted volume, respectively \citep{journal/MedIA/karim2018}.
Furthermore, \citet{journal/MedIA/veni2017} proposed to compare the scar percentages within the manually and automatically segmented LA wall. 
The basic idea behind this is that the LA wall segmentation is usually regarded as an initial step for the scar segmentation as mentioned earlier.

\subsubsection{LA scar measures}

\begin{figure}[t]\center
    \includegraphics[width=0.49\textwidth]{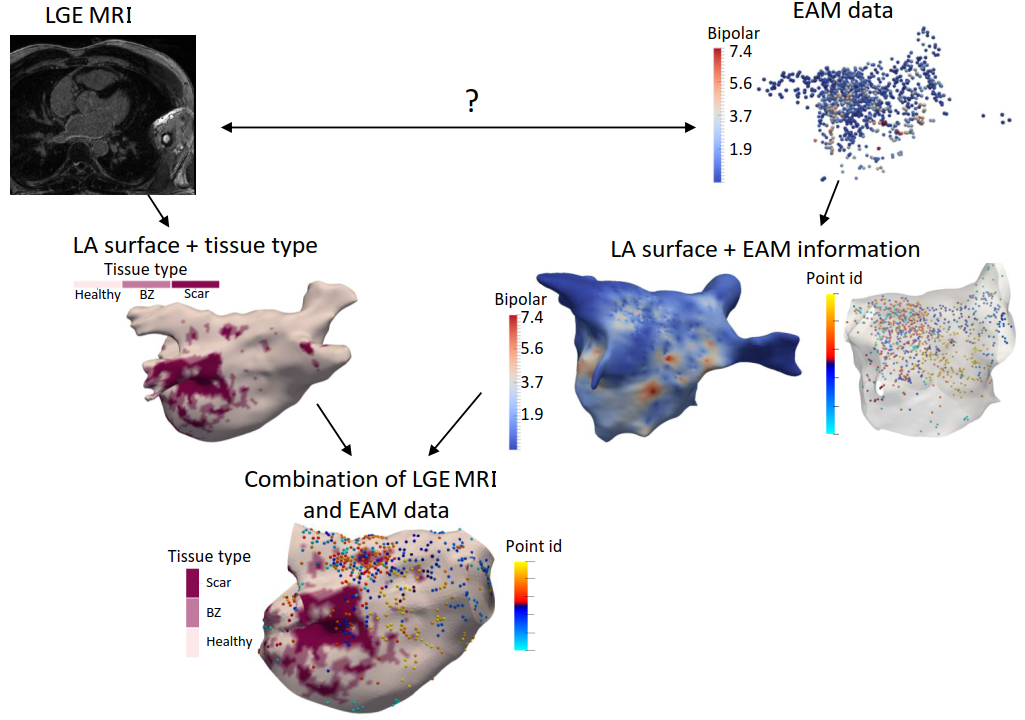}
   \caption{The spatial correspondence of LGE MRI and EAM data. Image adapted from \citet{thesis/UPF/nunez2018} \reviewerone{}{with permission}.}
\label{fig:evaluation:EAM}
\end{figure}

The optimal evaluation method to quantify scars from LGE MRI is still controversial due to the lack of ground truth.
Currently, the EAM system is regarded as the clinical standard technique for the scar assessment, as presented in \Leireffig{fig:evaluation:EAM}.
The widely used bipolar voltage threshold defining the LA scars is $\leq 0.05$ mV, which has been propagated through the literature and clinical practice \citep{journal/EHJ/harrison2014}.
However, the correlation between the LA scars identified by LGE MRI (enhanced regions) and EAM (low voltage regions) is still being questioned \citep{journal/Dia/floria2020}.
The subjective and inaccurate scar segmentation might be one of the main reasons.

Alternatively, most algorithms employ manual segmented LA scars as the ground truth.
For this evaluation, volume overlap measures and scar percentage are commonly used, as \Leireftb{tb:method:LA scar} shows.
For example, \citet{conf/MI/perry2012} proposed a novel overlap measure for the scar evaluation, namely XOR overlap,
\begin{equation}
  \mathrm{XOR}(V_{\mathrm{auto}}, V_{\mathrm{manual}}) = \frac{\left|W\right| + \left|V_{\mathrm{auto}} \oplus V_{\mathrm{manual}}\right|}{\left|W\right|},
\end{equation}
where $\left|W\right|$ is the set of voxels that belong to the LA wall, and $\oplus$ refers to exclusive OR.
The XOR overlap measure emphasizes the difference between overlapping scars, and will not be affected by the size of scars.

However, as mentioned in Section~\ref{wall_measure} volume overlap measures (such as Dice) could be highly sensitive to the mismatch of small structures (namely scars here), so in instances it will impose disproportionate penalties on the algorithm.
To mitigate the effect of the small size of scars, \citet{journal/MedIA/li2020} proposed to project the appearance of scars onto the LA surface for both ground truth and automatic segmentation results,
and then calculate the Dice scores of scars on the projected LA surface instead of on the 3D volume 
\citep{conf/MICCAI/wu2018,conf/STACOM/lilei2018,journal/MedIA/li2020,conf/MICCAI/li2020}.  
Furthermore, \citet{journal/MedIA/li2020,conf/MICCAI/li2020} computed the generalized Dice (GDice) of scars from the projected LA surface for a better interpretation.
GDice is defined as follows,
\begin{equation}
  \mathrm{GDice} = \frac {2\sum_{k=0}^{N_{k}-1}\left| {S}_{k}^{\textit{auto}} \cap {S}_{k}^{\textit{manual}}\right|} {\sum_{k=0}^{N_{k}-1}(\left| S_{k}^{\textit{auto}}\right| + \left|S_{k}^{\textit{manual}}\right|)},
\end{equation}
where $S_{k}^{\textit{auto}}$ and $S_{k}^{\textit{manual}}$ indicate the segmentation results of label $k$ from the automatic method and manual delineation on the LA surface, respectively, $N_{k}$ is the number of labels.
Here, $N_{k}=2$, where $k=1$ represents normal wall and $k=1$ refers to scarring regions.

\citet{journal/jcmr/Karim2013} proposed a surface-based metric, which employed MIP to calculate the distance error between the mesh vertex points on the LA surface.
The distance error is defined as the root mean squared error (RMSE), i.e.,
\begin{equation}
    \mathrm{RMSE}=\sqrt{\frac{1}{N} \sum_{i=1}^{N} d\left(v_{i}^{auto}, v_{i}^{manual}\right)^{2}},
\end{equation}
where $v_{i}^{auto}$ and $v_{i}^{manual}$ are the set of mesh vertices belonging to scars from the prediction and ground truth, respectively.
The major limitation of the surface based metric is that targets with a significant amount of $FP$ scars will have a low RMSE error. 
Nevertheless, it can be overcome by combining the surface measure with a volume-based index.

Scar percentage is directly related to clinical categorization of  AF patients, as presented in \Leireftb{tb:intro:utah grade}, and thus should be appropriate as an assessment measure. 
Besides, one could analyze the relationship of scar percentages between manually and automatic scar segmentations, to evaluate the performance of automatic scar segmentation.
For example, \citet{journal/MedIA/veni2017} quantified the scar percentage correlation using the mean square error (MSE) and R-square value.
Many works also calculate the volume error of scars for evaluation, which is defined as,
\begin{equation}
  \delta V=\left|V_{auto}-V_{manual}\right|.
\end{equation}
Statistical measurements related to scar classification could be employed for evaluation, including Acc, Sen, Spe, receiver operating characteristic (ROC) curve, and balanced error rate (BER).

\begin{table*} [t] \center
    \caption{\reviewerfour{}{Summary of representative results for LA LGE MRI computing on public AF-related datasets.
    CoV: coefficient of variation; CV: conventional methods.} }
\label{tb:result:summary}
{\small
\begin{tabular}{l|ll}
\hline
Public dataset source    & Target & Representative result \\ 
\hline
\citet{link/CARMA2012}          & LA cavity             & \tabincell{l}{Dice = $0.79 \pm 0.05$, OV =  $0.65 \pm 0.07$, ASD = $2.79 \pm 2.84$ mm, 95\% HD = $14.4 \pm 3.65$ mm \\ \citep{journal/TIP/zhu2013}}  \\ 
\hdashline   
\multirow{6}*{\citet{journal/MedIA/xiong2020}}  & \multirow{6}*{LA cavity}  & Dice = $0.861 \pm 0.036$ (CV) \citep{conf/STACOM/qiao2018}, $0.942 \pm 0.014$ (DL) \citep{journal/TMI/xiong2018} \\ 
~&   ~& Jc = $0.758$ (CV) \citep{conf/STACOM/qiao2018}, $0.874$ (DL) \citep{conf/STACOM/xia2018} \\
~&   ~& HD = $11.8$ mm (CV) \citep{conf/STACOM/qiao2018}, $8.60$ mm (DL) \citep{conf/STACOM/chen2018} \\
~&   ~& ASD = $1.473$ mm (CV) \citep{conf/STACOM/nunez2018}, $0.748$ mm (DL) \citep{conf/STACOM/xia2018} \\
~&   ~& Sen = $0.847$ (CV) \citep{conf/STACOM/qiao2018}, $0.949$ (DL) \citep{conf/STACOM/preetha2018} \\  
~&   ~& Spe = $0.999$ (CV) \citep{conf/STACOM/qiao2018}, $1.000$ (DL) \citep{conf/STACOM/vesal2018} \\   
\hdashline                                               
\citet{journal/MedIA/zhuang2019}                & \tabincell{l}{Whole heart includ- \\ing LA cavity} & Dice = $0.844 \pm 0.097$ (only on AF patients) \\
\hdashline
\multirow{3}*{\citet{journal/MedIA/karim2018}}  & \multirow{3}*{LA wall}          & \tabincell{l}{Dice = 0.43 (A) \citep{conf/STACOM/tao2016}, 0.39 (P)           \citep{conf/STACOM/inoue2016}, $0.67 \pm 0.22$ \\  (Inter-ob) on the CT; $0.72$, $0.56 \pm 0.14$ (Inter-ob) on the MRI} \\
~&   ~& Tk error = 0 (P/A), 0.25 mm (P) and 0.20 mm (A) (Inter-ob) \\  
~&   ~& Tissue mass error = 3.84$\sim$14.63 g, $10.03 \pm 4.0$ g (Inter-ob)\\  
\hdashline                                               
\multirow{3}*{\citet{journal/jcmr/Karim2013}}   & \multirow{3}*{LA scar}      & Dice = $0.85^{post}$, $0.48^{pre}$  \\
~&  ~& RMSE = $6.34 \pm 8.2^{post}$ mm, $0.17 \pm 0.1^{pre}$ mm \citep{journal/QIMS/lu2012} \\
~&  ~& $\delta V$ = $6.34 \pm 8.2^{post}$ ml, $1.25 \pm 1.5^{pre}$ ml \\
\hdashline 
\citet{link/CARMA2012}                          & LA scar               & \tabincell{l}{Dice = $0.807 \pm 0.106$ (semi-auto) \citep{conf/MI/perry2012} \\ XOR = $0.916 \pm 0.035$ (semi-auto) \citep{conf/MI/perry2012} \\ CoV = 0.62 \citep{conf/CinC/andalo2018}} \\
\hline
\end{tabular} }\\
\end{table*}

\subsubsection{LA ablation gap measures}
As \Leireftb{tb:method:LA gap} shows, most gap quantification methods in the literature employed ablation gap characteristics (i.e., number, length, and position of gaps) for evaluation.
Similar to the evaluation of scars, these works also analyzed the correlation with EAM, by comparing the ablation gaps in LGE MRI to the electrical gaps in EAM.
However, the applicability of EAM for ablation gap quantification is limited.
This is mainly because: 
1) the difficulty of the gap position registration between LGE MRI and EAM; 
2) the voltage mapping does not entirely reflect scar/ gap formation; 
3) the requirement of a voltage threshold for scar/ gap classification, with the same issues as for the LGE MRI threshold. 
Therefore, direct extrapolation of EAM data to verify LGE MRI should be performed carefully, in particular when they offer contradictory information \citep{journal/MedIA/nunez2019}.
Besides, \citet{journal/CAR/ranjan2012} calculated the correlation between gap length (GL) measured via LGE MRI for evaluation.
\citet{journal/MedIA/nunez2019} proposed a quantitative index, i.e., relative gap measure (RGM), to calculate the proportion of the ablation gaps on a defined standard LA parcellation.
\begin{equation}
  \mathrm{RGM} = \frac{\text {Gap length}}{\text {Encircling Path length}},
\end{equation}
where ``Gap length" indicates the sum of all GLs along the ``Encircling Path", and the ``Encircling Path length" refers to the length of the complete closed-loop on the PVs. 
The RGM is between 0 and 1, which means that if RGM = 0, the vein is completely surrounded, and if RGM = 1, there are no scars around the veins.
To alleviate the effect of the scar segmentation, one could adopt a multi-threshold scheme for the scar segmentation, and then integrate the results into the RGM calculation \citep{journal/MedIA/nunez2019}.

\subsection{\reviewerfour{}{Evaluation results on the AF-related public dataset}} \label{evaluation results}

\reviewerfour{}{In general, the segmentation accuracy of different methods is not directly comparable, unless these methods are evaluated on the same dataset using the protocols. 
Therefore, we only summarize the state-of-the-art results of reviewed LA LGE MRI computing methods on the public dataset here, as presented in \Leireftb{tb:result:summary}.}

\reviewerfour{}{For LA cavity segmentation, three public datasets are available, and Dice, ASD and HD are commonly used for evaluation.
On the dataset from \citet{link/CARMA2012}, the state-of-the-art results of the LA cavity segmentation in all metrics were from \citet{journal/TIP/zhu2013}.
On the dataset from \citet{journal/MedIA/zhuang2019}, mean Dice scores from different methods have been reported for each pathology including AF.
The methods evaluated on the public dataset from \citet{journal/MedIA/xiong2020} have been separated into conventional methods and DL-based methods.
For each metric, we list the state-of-the-art results from conventional and DL-based methods, and the best Dice score for the LA cavity segmentation was obtained by \citet{journal/TMI/xiong2018} (Dice = $0.942 \pm 0.014$).
The DL-based methods demonstrated great potential, as the best result in each metric was all obtained by DL-based methods on this dataset.}

\reviewerfour{}{For LA wall segmentation, there is only one available public dataset for evaluation, which included 10 CT and 10 non-enhanced MRIs instead of the LGE MRI.
We present both the state-of-the-art results and inter-observer variations for each metric.
One can see that the results based on semi-automatic algorithms were generally comparable to the inter-observer variations for each metric.
However, the size of this dataset is small, and current semi-automatic methods are labor-intensive and subjective.}

\reviewerfour{}{For LA scar segmentation, two public datasets are accessible, and typically Dice is used for evaluation.
On both datasets, only semi-automatic algorithms were applied.
There was performance variation among pre- and post-ablation images from \citet{journal/jcmr/Karim2013}.
Specifically, the best Dice scores were 0.48 and 0.85 on pre- and post-ablation LGE MRIs, respectively.
However, in terms of RMSE and $\delta V$, the performance on the pre-ablation LGE MRIs was better than that on the post-ablation LGE MRI.
The possible reason could be that the volume of post-ablation LGE MRI is generally larger than that of pre-ablation image.
Nevertheless, pre-ablation LGE MRI is still generally more challenging for fibrosis segmentation due to its more diffuse distributions.
}

\section{Potential clinical applications of the developed algorithms} \label{clinical aplication}

It is essential to evaluate the clinical utility of the developed approaches for AF.
Instead of blindly improving the accuracy of methods, researchers therefore can focus more on answering some clinical questions related to AF.
The exploration and understanding of potential clinical applications of AF can guide the development of segmentation and quantification algorithms and answer important clinical questions. 
\reviewerfour{}{For example, we can employ the developed segmentation and quantification techniques to compare native and ablation-induced scars (Section \ref{discussion:pre_post}), inspect the regional distribution of wall thickness (Section \ref{discussion:region distribution}), fibrosis/ scars and ablation gaps from LGE MRI, and analyze the relationship between fibrosis/ scars/ gaps and AF recurrence (Section \ref{discussion:relationship analysis}).
Moreover, there are several other clinical applications, such as analyzing the relationship between the low-voltage regions in EAM and scars detected by LGE MRI, the relationship between ablation parameters (power of the radiofrequency signal, catheter contact force, etc.) and the created chronic lesion detected by LGE MRI, as well as assessing the reproducibility of LGE MRI scar imaging with respect to imaging parameters.
However, the latter three applications require additional EAM data or LGE MRIs with different ablation and imaging parameters, and therefore are out of the scope of this review.}

\reviewertwo{}{To the best of our knowledge, there are a limited number of review papers targeting the clinical applications of LGE MRI. 
\citet{journal/CJC/zghaib2018} summarized the new insights into the use of MRIs for the decision-making of AF management. 
They explored LGE, native T1-weighed, T2-weighted as well as cine MRI, and for LGE MRI they only reviewed studies on the relationship between the extent of scars on post-ablation LGE MRIs and the rate of AF recurrence. 
In this section, we will provide a comprehensive review from the perspective of the clinical applications for AF analysis.}


\subsection{Comparisons of native and ablation-induced scars} \label{discussion:pre_post}

Recent studies demonstrated the differences in the extent and distribution of fibrosis/ scars of pre-/ post-ablation LGE MRI \citep{journal/HR/malcolme2013,journal/HR/fukumoto2015}.
For instance, \citet{journal/HR/malcolme2013} found that there was no difference of scars between ostial and LA cavity regions for pre-ablation data, but in post-ablation data the extent of scars in the ostia is larger than that in the LA cavity.  
They also reported a positive association between the extent of preexisting fibrosis and AF recurrence, which coincides with the finding in the literature \citep{journal/JACC/verma2005,journal/HR/mahnkopf2010}.
However, they did not find any relationship between the amount of ablation-induced scars and AF recurrence, which should be negatively associated according to the studies of \citet{journal/JACC/peters2009,journal/JACC/mcgann2011}.
\citet{journal/HR/fukumoto2015} demonstrated that ablation-induced scars are related to greater contrast affinity and thinner walls compared to preexisting fibrosis.
\citet{conf/MI/yang2017} tried to distinguish native and ablation-induced scars via a texture based feature extraction.
They stated the difficulty of the differentiation between native and ablation-induced scars, especially for longstanding persistent AF.
Therefore, the understanding of the characteristics of pre- vs. post-ablation scars can be important and may inform future ablation strategies for AF.

\begin{figure*}[t]\center
    \subfigure[] {\includegraphics[width=0.65\textwidth]{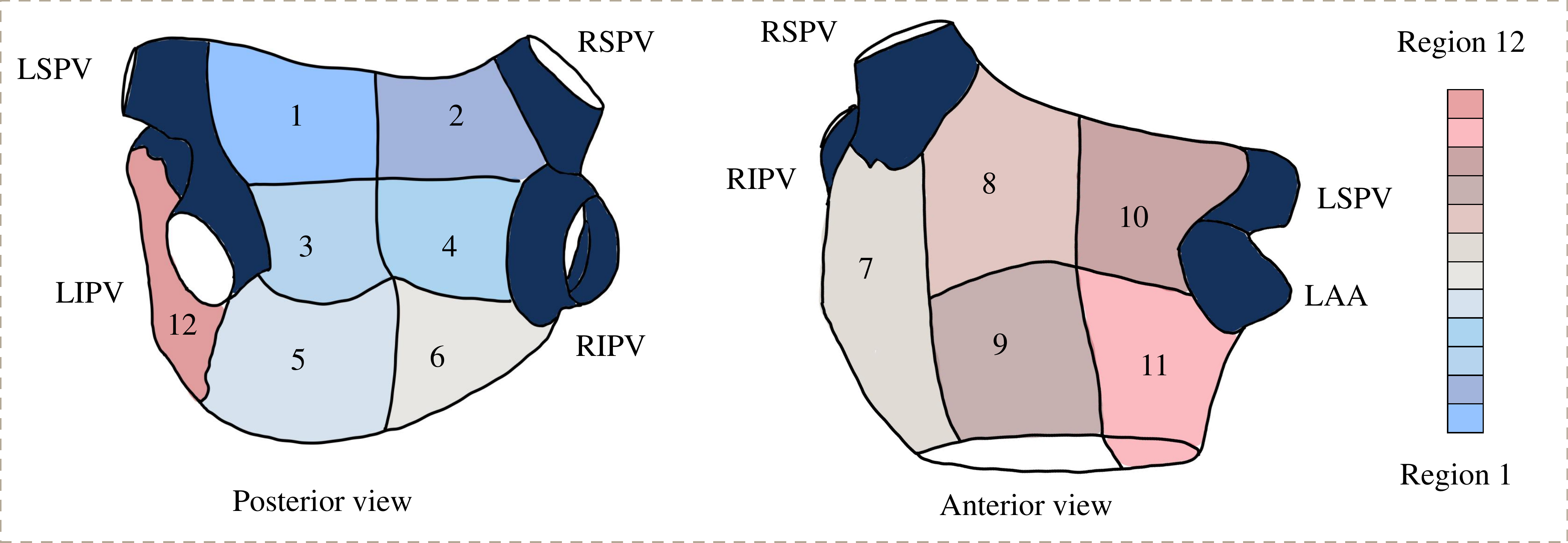}}
    \subfigure[] {\includegraphics[width=0.32\textwidth]{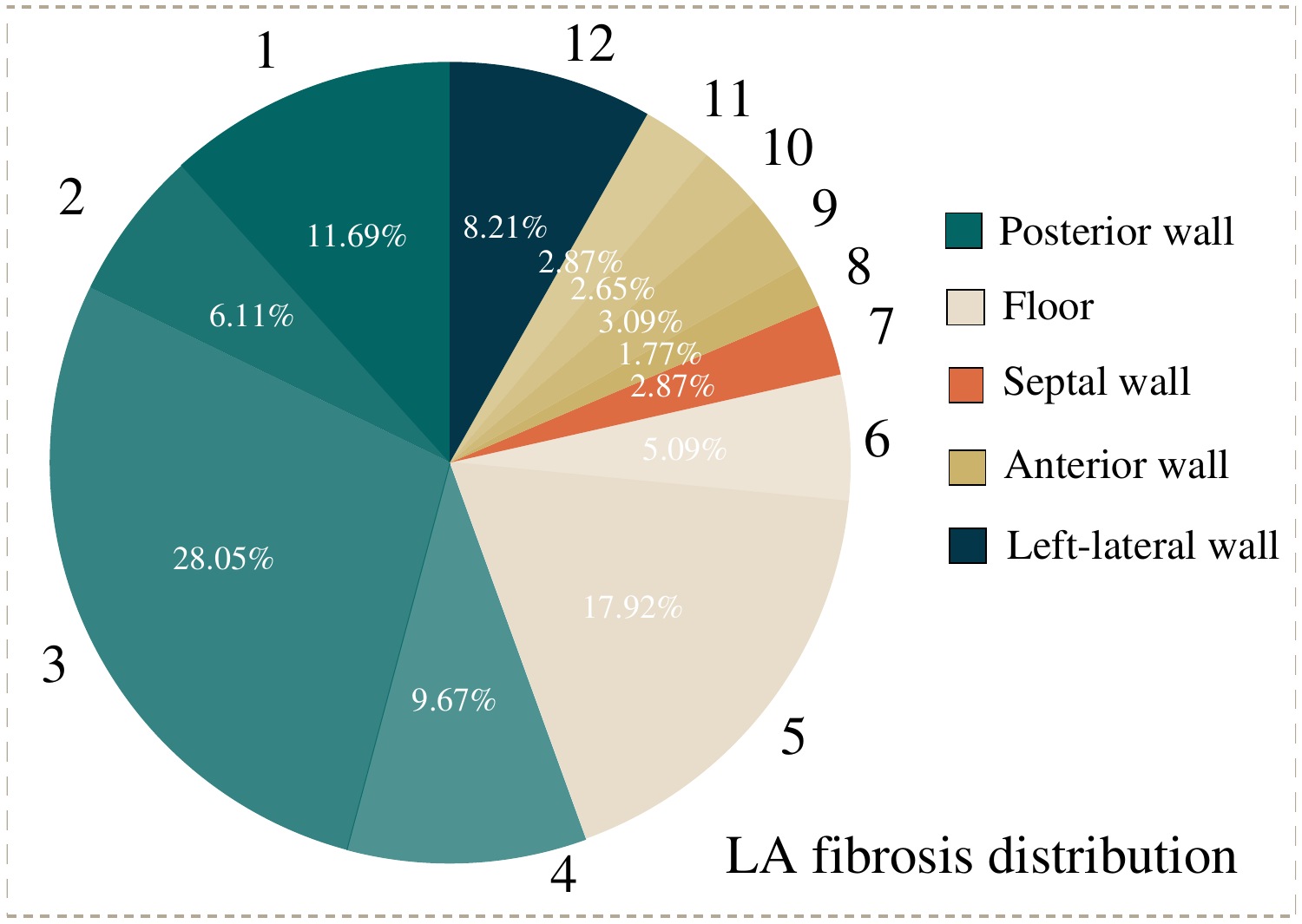}}
   \caption{\reviewerone{}{Example of a LA parcellation and its corresponding fibrosis distribution: 
   (a) the LA surface template parcellated from anatomical landmarks; 
   (b) the regional distribution of LA fibrosis (the same color with different transparencies refers to the same region category), where the values were obtained from \citet{journal/Ep/benito2018}. 
   Illustrations designed referring to \citet{journal/Ep/benito2018}.}}
\label{fig:discussion:scar_distribution}
\end{figure*}

\subsection{Regional distribution analysis of wall thickness and fibrosis/ scars} \label{discussion:region distribution}
To date, there are already several studies on LA wall thickness measurements, to analyze the relationships between wall thickness and patient age, AF stage/ type, scar formation, and AF recurrence \citep{journal/MedIA/karim2018}.
For example, \citet{journal/JICE/hall2006} studied 34 patients of different ages and found that the thinnest and thickest areas were the roof ($1.06 \pm 1.49$ mm) and septum ($2.2 \pm 0.82$ mm), respectively. 
They did not find any significant relationships between the wall thickness and age.
In contrast, \citet{journal/Chest/pan2008} measured the wall thickness on 180 AF patients of various ages and concluded that the thickness increased with age.
They also found that the anterior wall ($2.0 \pm 0.9$ mm, $3.2 \pm 0.2$ mm and $3.7 \pm 0.9$ mm in 40$\sim$60, 60$\sim$80 and 80+ year olds) was thicker than the posterior wall ($0.7 \pm 0.2$ mm, $1.8 \pm 0.2$ mm and $2.4 \pm 0.4$ mm in 40$\sim$60, 60$\sim$80 and 80+ year olds) among all the age groups.
\citet{journal/JCE/beinart2011} and \citet{journal/JICE/hayashi2014} both observed that the middle superior posterior wall was the thinnest region with a thickness of $1.43 \pm 0.44$ mm and $1.44 \pm 0.17$ mm, respectively.
\citet{journal/HV/suenari2013} analyzed the thickness of 54 AF patients, and showed that the thickest wall area is in the left lateral ridge ($4.42 \pm 1.28$ mm), while the thinnest is in the LIPV ($1.68 \pm 0.27$ mm).
Besides, they found that the thickness of the left lateral ridge was correlated to the AF recurrence ($p$=0.041).
However, the superior right posterior wall was found to be significantly associated with both AF recurrence ($p$=0.048) and electrical reconnection ($p$=0.014) in \citet{conf/JCE/inoue2016}.
Despite this progress, most of these works were based on manual segmented the LA wall, and focused on CT images instead of LGE MRI.
Note that transmural lesion formation is critical to the success of AF ablation and is dependent on the knowledge of regional LA wall thickness.
Therefore, the distribution analysis of wall thickness from LGE MRI could be important and might provide insight into the progress of the AF. 

As for the regional distribution of fibrosis/ scars in the LA LGE MRI, related information is limited and has not been comprehensively reported.
\citet{journal/JCE/cochet2015} divided the LA into four segments and reported an irregular fibrosis anatomical distribution.
However, they found that fibrosis generally occurred more often on the posterior LA wall than the anterior one, particularly in the area adjacent to and below LIPV.
\citet{journal/Ep/benito2018} manually defined the LA parcellation with 12 sub-regions: 1$\sim$4, posterior wall; 5$\sim$6, floor; 7, septal wall; 8$\sim$11, anterior wall; 12, lateral wall (see \Leireffig{fig:discussion:scar_distribution} (a)).
They selected 76 consecutive AF patients for analysis and also observed that the fibrosis was preferentially located at the posterior wall and floor around the antrum of the LIPV, i.e., segments 3 and 5 (40.42\% and 25.82\% fibrosis), as \Leireffig{fig:discussion:scar_distribution} (b) shows.
In contrast, segments 8 and 10 (2.54\% and 3.82\% fibrosis) in the anterior wall contained the fewest fibrosis.
Similar to the increased wall thickness in \citet{journal/Chest/pan2008}, they found that age (\textgreater 60 years old) was also significantly correlated to increased fibrosis ($p$=0.04).
Recently, \citep{journal/Radiology/lee2019} separated the LA into nine segments, and also found that scars were most frequently seen at the posterior wall around the LIPV.
Besides, they studied 195 paroxysmal and 121 persistent AF patients and observed that the presence of fibrosis assessed in LIPV from LGE MRI was associated with the chronicity of AF.
This preliminary research suggests that the knowledge of preferential fibrosis/ scar position may open further perspectives in ablation strategies, patient selection, and AF recurrence prediction.

\subsection{Relationship analysis between fibrosis/ scars/ gaps and AF recurrence} \label{discussion:relationship analysis}
As mentioned in Section \ref{discussion:pre_post}, both the extent of preexisting fibrosis and ablation-induced scars are correlated with AF recurrence, but with opposite effects.
Specifically, AF recurrence is positively associated with the extent of preexisting scars, but negatively related to that of post-ablation scars.
The characteristics of pre- vs. post-ablation scars may explain the seemingly paradox and inform future strategies for ablation \citep{journal/HR/fukumoto2015}.
With respect to the pre-ablation scars (also namely fibrosis), it has been regarded as a potential cause of the abnormalities in atrial activation, which may underlie the initiation and maintenance of AF.
Note that AF belongs to a progressive disease, and several studies revealed that causality between AF and fibrosis may be bidirectional \citep{journal/Circulation/oakes2009}.
This might explain why patients with a greater extent of fibrosis normally suffer much higher recurrence rates after ablation.
Apart from the extent of fibrosis, \citet{journal/Circulation/oakes2009} investigated 81 AF patients with pre-ablation LGE MRI, and found that AF recurrence was also related to the locations of fibrosis.
In their experiments, patients with recurrent AF presented fibrosis on the whole LA, whereas patients without recurrent AF had fibrosis only located primarily to the posterior wall and septum.
As for post-ablation scars, robust evidence supports that complete circumferential and transmural lesion formation is critical to successful AF ablation \citep{journal/Circulation/cappato2003,journal/Circulation/verma2005,journal/Circulation/ouyang2005}.
Here, the ablation lesion just refers to the post-ablation scars or can be named ablation-induced scars.
Therefore, patients with a smaller degree of post-ablation scars on LGE MRI tend to recur AF after ablation.
Similar to fibrosis, the location of post-ablation scars is also an important index for AF recurrence prediction.
For example, several studies emphasized the importance of right inferior PV (RIPV) scars, which is the most highly correlated to clinical ablation success \citep{journal/Europace/yamada2006,journal/JACC/peters2009}.
This could attribute to the reported technical difficulty in ablating the RIPV region due to poor catheter access, resulting in its greater variability of scars. 
For example, \citet{journal/JACC/peters2009} studied 35 AF patients undergoing the first ablation procedure, and compared the extent of scars on different sub-regions.
They demonstrated that the PVs of patients without recurrence had more completely circumferential scars, especially on RIPV regions.
In the case of ablation gaps, which are generally caused by incomplete PVI, the extent and distribution of gaps are regarded to be positively associated with AF recurrence.
The identification and localization of ablation gaps from LGE MRI have been used to predict AF recurrence and further guide repeated PVI procedures \citep{journal/JACC/bisbal2014}.



\section{Discussion and future perspectives} \label{discussion}
\reviewertwo{}{LGE MRI has attracted increasing attention in the assessment of AF before and after an ablation procedure. 
Automatic segmentation and quantiﬁcation algorithms of LA structures and tissues can facilitate the diagnosis and therapy of AF patients. 
However, the translation of current algorithms into the clinical environment remains challenging. 
In this section, we summarize existing major challenges in the ﬁeld of LA LGE MRI computing and the solutions recently proposed. 
The exploration of these challenges and related works is expected to provide useful information for developing novel methods and applications for AF analysis.
}


\subsection{\reviewertwo{}{Surface projection and LA unfolding mapping}}

\begin{figure*}[t]\center
    \subfigure[] {\includegraphics[width=0.62\textwidth]{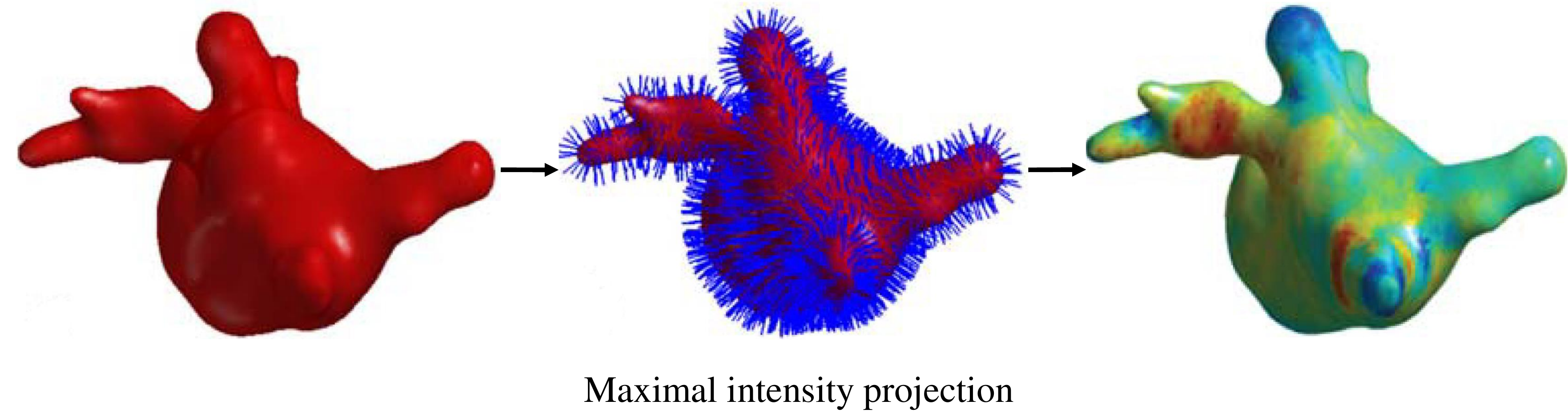}}
    \subfigure[] {\includegraphics[width=0.34\textwidth]{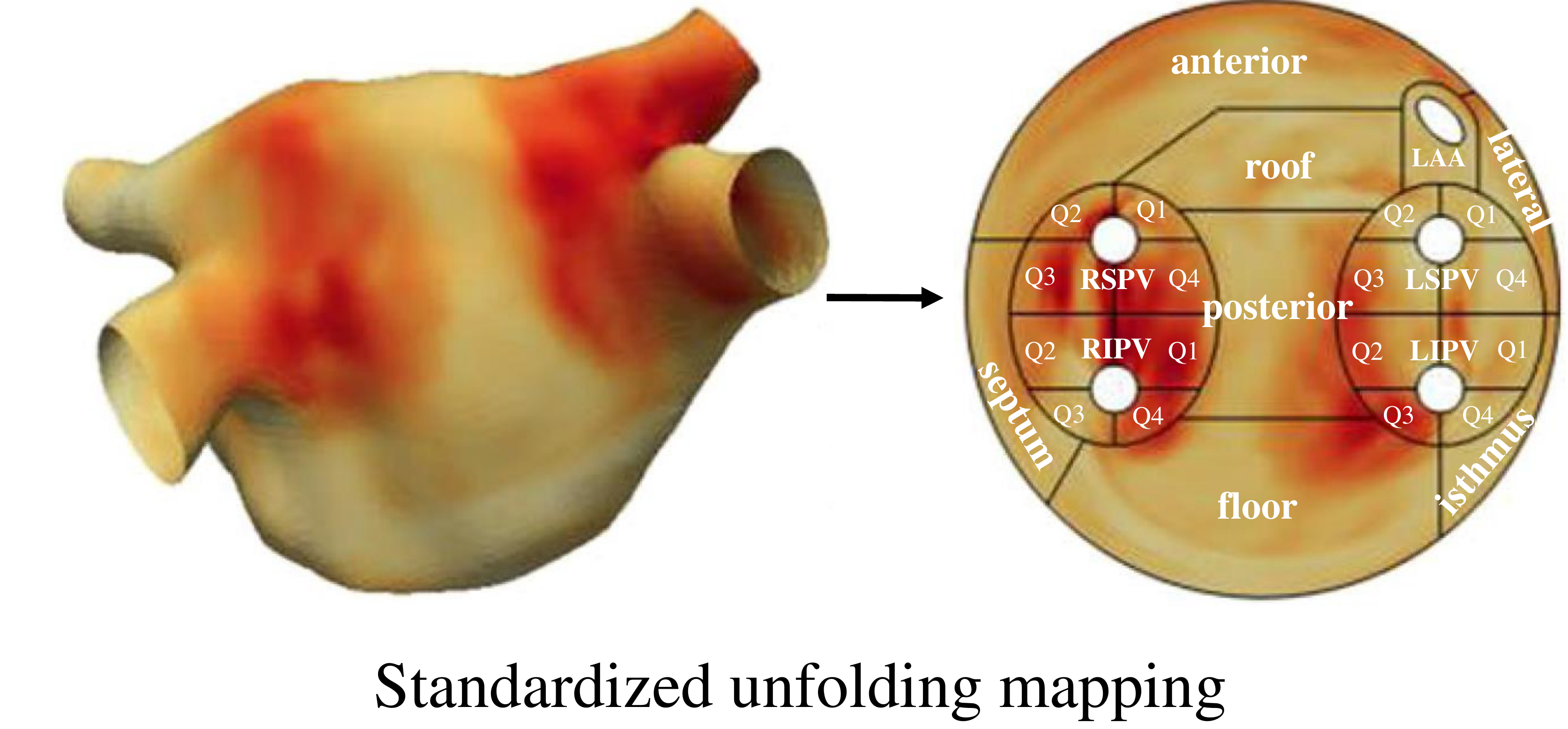}}
   \caption{\reviewertwo{}{Alternative visualizations and representations of LA scars via: 
   (a) maximum intensity projection (images adapted from \citet{journal/JMRI/tao2016} with permission); 
   (b) LA standardized unfolding mapping (images adapted from \citet{journal/JICE/williams2017} with permission).}
   }
\label{fig:discussion:projection}
\end{figure*}

\reviewertwo{}{Recent studies have shown that the success of AF treatment highly relies on the formation of contiguous and transmural scars on the LA wall \citep{journal/JACC/glover2018}.
However, the wall thickness is difficult to measure based on current LGE MRI techniques.
In clinical practice, the location and extent of scars are believed to have greater clinical significance and can be used to predict outcomes of AF ablation procedures \citep{journal/CAE/arujuna2012}.
Therefore, several studies have been proposed to project scars onto the LA surface to perform scar quantification \citep{journal/TMI/ravanelli2014,journal/JMRI/tao2016,journal/MedIA/li2020,journal/MedIA/li2021}.
\Leireffig{fig:discussion:projection} (a) presents an example of scar projection achieved by MIP.
By projection, the errors due to LA wall thickness can be mitigated, and the computational complexity of algorithms can be drastically reduced.}

\reviewertwo{}{Nevertheless, the cross-subject comparison of 3D surface data is still arduous.
To solve this, \citep{journal/MedIA/roney2019} developed a universal atrial coordinate mapping system for 2D visualization of both the LA and right atrium.
\citet{journal/JICE/williams2017} created a 2D LA standardized unfolding mapping (LA-SUM) template where the MV was mapped to a disk, the PVs to circles, and the LAA to an ellipse, as presented in \Leireffig{fig:discussion:projection} (b).
The target 3D LA will be registered to a 3D template and then transferred to the 2D template via a 3D-2D template mapping.
The LA flattening of LA-SUM may result in undesired information loss between 3D and 2D LA representations due to the possible inaccurate registration between LA surfaces with high shape variability. 
Instead of relying on a 3D registration step, \citep{journal/TVCG/nunez2020} proposed a quasi-conformal LA flatting scheme and employed additional regional constraints to overcome undesired mesh self-folding.
The advantages of these LA unfolding mapping techniques include 2D visualization, LA regional assessment, and multi-modal data combination.
However, their templates were generally designed for the most common LA topology with four PVs.
We therefore expect that more flexible templates can be developed to adapt for the LA topological variants.}

\subsection{\reviewertwo{}{Joint optimization and independent analysis of the AF-related tasks}}
\reviewertwo{}{The target regions of the four tasks reviewed in Section \ref{method} are all inherently related, particularly in the spatial information of images, as shown in \Leireffig{fig:intro:review_structure}.
Several studies employed multi-task learning for simultaneous LA cavity segmentation and scar segmentation/ quantification and proved the effectiveness of joint optimization \citep{conf/MICCAI/chen2018,conf/MICCAI/li2020}.
The spatial information between the LA cavity and scars could simply be learned via spatial attention, i.e., multiplying the LA cavity feature map by the scar feature map \citep{conf/MICCAI/chen2018}, or projecting the scars onto the LA endocardial surface \citep{conf/MICCAI/li2020}.
At the same time, several studies have been devoted to reducing the correlation between the accuracy of related tasks in LA LGE MRI computing, i.e., their conditional dependencies.
For instance, MIP schemes have been widely used in LA scar and gap quantification to mitigate the effect of inaccurate LA cavity segmentation \citep{journal/TBME/knowles2010,journal/JMRI/tao2016,conf/MI/razeghi2020,journal/JACC/bisbal2014}.
Patch shift scheme was developed to apply a random shift along the LA boundary when performing surface projection \citep{journal/MedIA/li2020}.
\citet{conf/MICCAI/li2020} learned the spatial information around the LA boundary to reduce the dependence on accurate LA cavity segmentation.
Despite these advances, the joint optimization and independence analysis of the AF-related tasks are yet to be explored in further depth in the future.
}

\subsection{\reviewerthree{}{Challenges with deep learning in LA LGE MRI computing}} \label{discussion:DL in LA computing}
\reviewerthree{}{It is evident that DL-based methods have obtained promising results on the LA cavity and scar segmentation and quantification.
It is mainly attributed to the release of related public datasets and the emerge of advanced network architectures.
\reviewertwo{}{With the release of public datasets, the research on the LA cavity and scar segmentation from LGE MRI started to increase, as \Leireffig{fig:intro:review_distribution} shows.}
Despite the promising results, deep neural networks still confront a number of challenges, such as poor interpretability, scarcity of annotated data, class imbalance problems, limited domain generalization ability, and catastrophic forgetting.
One may refer to the review papers \citep{journal/FCM/chen2020,journal/JDI/hesamian2019} to follow these challenges and state-of-the-art solutions for DL-based medical image segmentation.
Here, we mainly discuss the limited data (Section \ref{DL: limited data}) and model generalization issues (Section \ref{DL: limited generalization}), as there exist several unique points in the two challenges for AF studies.}

\subsubsection{\reviewerthree{}{Scarcity of (annotated) data}} \label{DL: limited data}
\reviewerthree{}{The scarcity of (annotated) data is a serious issue in LA LGE MRI computing.
Though this is common in many other tasks, LGE imaging could be more challenging, due to the existence of contrast enhancement, its complex patterns, and the large quality and contrast variations across different patients. 
Especially, LGE MRI of LA wall requires substantially higher spatial resolution, patient-specific optimization of scan parameters, strict criteria for contrast dosage and delay between contrast injection and image acquisition, compared to LGE MRI of the LV \citep{journal/JACC/siebermair2017,journal/JCMR/chubb2018}.
These precise requirements are difficult to meet in practice, resulting in scarcity and poor image quality of LGE MRI.
It is also complicated to collect many annotated cases of 3D LGE MRI.
However, DL-based LA LGE MRI computing typically relies on a large number of annotated samples for training.
Several schemes have been proposed to solve this.
For example, \citet{conf/MICCAI/yu2019} employed a semi-supervised learning method for the LA cavity segmentation from LGE MRI, to fully utilize the unlabeled data.
\citet{journal/MedIA/li2020} adopted a patch-wise training for the LA scar quantification from LGE MRI, which considerably increased the amount of labeled training data.
Data argumentation is generally useful in deep learning with limited training data, for example the method of partially region rotation of scars was employed for LV segmentation from LGE MRI \citep{conf/STACOM/campello2019}. 
Unsupervised domain adaptation has also been proven to be capable to alleviate the problem of limited annotated data from the target domain, which has been widely used for LV LGE MRI segmentation \citep{journal/MedIA/zhuang2020,journal/TMI/wu2020,journal/TMI/wu2021,journal/MedIA/pei2021}.
Finally, the methods making full use of sparse annotation \citep{conf/MICCAI/cciccek2016} are promising for LA LGE MRI computing with limited annotated data and could be further explored in the future.}

\subsubsection{\reviewerthree{}{Limited domain generalization ability}}  \label{DL: limited generalization}
\begin{figure}[t]\center
    \includegraphics[width=0.48\textwidth]{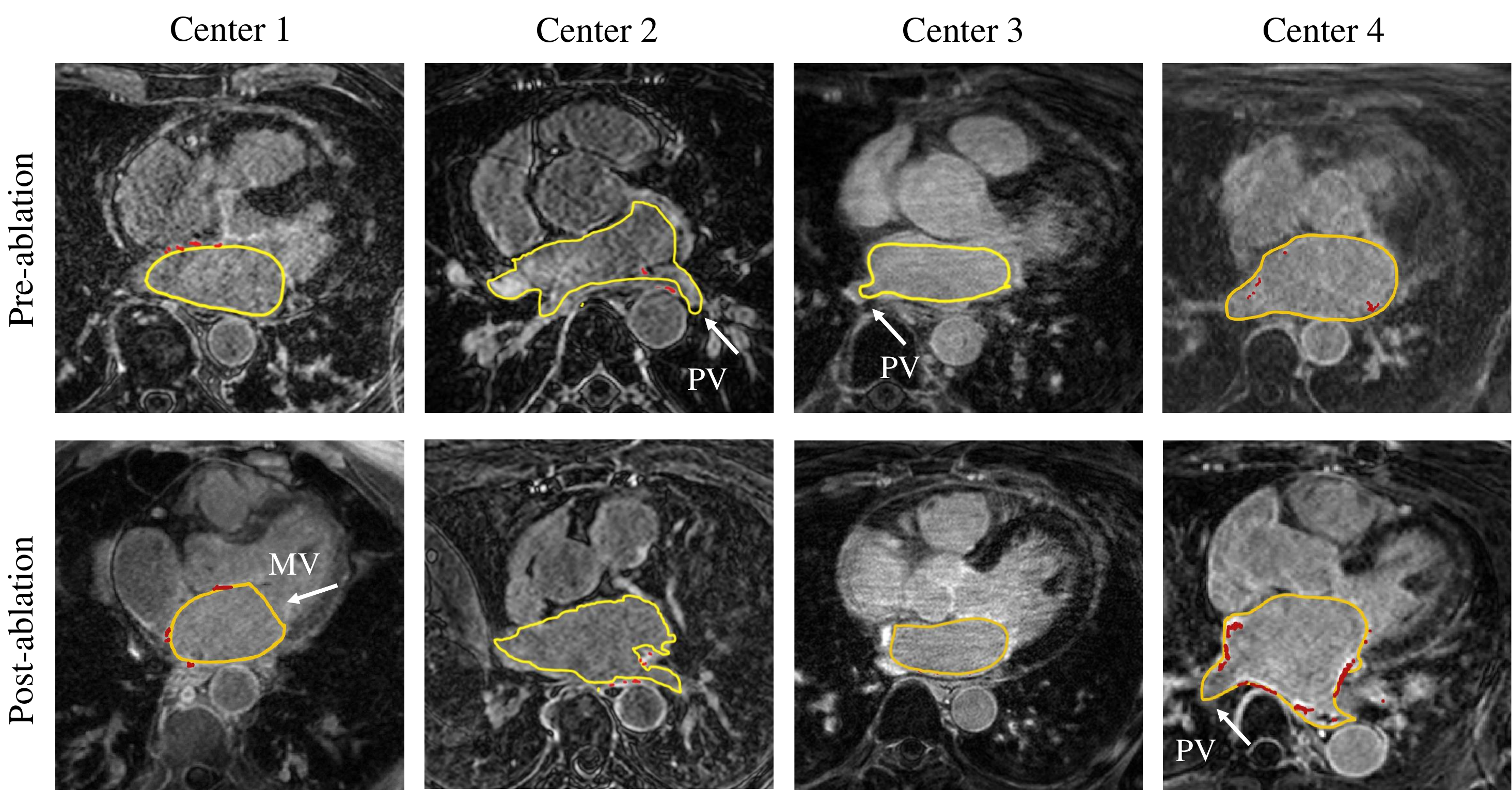}
   \caption{Multi-center pre- and post-ablation LGE MRIs, where LA cavity contour is labeled in the yellow and scarring region is labeled in red. 
   The images differ in contrast, enhancement as well as background, and the labels across different centers also exist variations, especially in the MV and PV regions.}
\label{fig:discussion:multi-center LGE MRI}
\end{figure}

\reviewerthree{}{Currently, most existing algorithms have only been evaluated on center- and vendor-specific LGE MRI.
Though the \textit{Left Atrium Fibrosis and Scar Segmentation Challenge} offered multi-center and multi-scanner data, the benchmark algorithms only tested on center- and vendor-specific images. 
Their suitability and performance had not been tested on data from other centers or vendors \citep{journal/jcmr/Karim2013}.
Note that LGE MRIs from different centers can vary evidently in appearance, as \Leireffig{fig:discussion:multi-center LGE MRI} shows.
This is mainly due to the absence of standardized LGE MRI acquisition protocols, leading to poor reproducibility of LGE MRI \citep{journal/Ep/benito2017,journal/JACC/sim2019}.
Even in the same dataset, one could encounter a severe data mismatch problem, resulting in poor outlier results \citep{journal/MedIA/li2021}.
Several schemes have been employed to solve this, such as data augmentation/ generation, domain-invariant representation learning, and meta-learning \citep{journal/arXiv/wang2021}.
Nevertheless, large multi-center and multi-scanner datasets are needed to validate the robustness and generalizability of current methods, which is more useful in practice.
It is also worthy of promoting deep models with efficient inherent generalization abilities for the LGE MRI data processing from different centers and vendors \citep{conf/MICCAI/li2021}.
Moreover, it could be interesting to study the domain shift between pre- and post-ablation LGE MRIs from the same center, and the label variations of LGE MRIs from different centers.}

\section{Conclusion} \label{conclusion}
We have presented and discussed the current progress of LGE MRI computing for LA studies, particularly for the four tasks, including segmentation and (or) quantification of LA cavity, wall, scars, and ablation gaps.
Though LGE MRI has been proven to be a powerful diagnostic and prognostic tool in the study of AF, a standardized imaging protocol should be further investigated.
Furthermore, a limited number of works have been reported focusing on image computing tasks, especially for automatic LA wall segmentation and ablation gap quantification.
Most research relies on manual delineation for further analysis and clinical applications.
Therefore, more accurate and robust automatic methods are desired for overall wide and intelligent use in the clinical setting.
The data-driven approaches have shown great potential for the LA cavity and scar segmentation and quantification, thanks to the development of deep neural networks.
The joint optimization of these related tasks can be a new direction for the utilization of their spatial relationship.
To research for a broader clinical application, well-controlled and large-cohort studies are expected to better guarantee the reproducibility of measurements, refine the evaluation methods, and validate the impact on clinical outcomes as well as the computing accuracy. 

\reviewertwo{}{
Although we limit our survey related to AF analysis in the article, the described methodologies can be useful to other clinical applications. 
We described in detail the characteristics of targets, which motivated the methodologies. 
Consequently, such methods can be used for other targets sharing similar characteristics as the targets in AF studies. 
For instance, tumor lesions are also small and diffuse targets, so the review on the scar segmentation and quantification methods could inspire the development of methods on tumor lesion segmentation, and vice versa. 
We believe that this review has the potential to help researchers to design appropriate frameworks according to their problems and be aware of similar challenging issues and state-of-the-art solutions. 
}


\section*{Acknowledgment}
This work was supported by the National Natural Science Foundation of China (61971142, 62111530195 and 62011540404) and the development fund for Shanghai talents (2020015).
L Li was partially supported by the CSC Scholarship.
JA Schnabel and VA Zimmer would like to acknowledge funding from a Wellcome Trust IEH Award (WT 102431), an EPSRC programme grant (EP/P001009/1), and the Wellcome/EPSRC Center for Medical Engineering (WT 203148/Z/16/Z).

\bibliographystyle{model2-names}
\biboptions{authoryear}
\bibliography{A_refs}

\end{document}